\documentclass[sigconf]{acmart}
\AtBeginDocument{%
  }
\usepackage{algorithm}
\usepackage{algorithmic}
\usepackage[table]{xcolor}
\usepackage{booktabs}   
\usepackage{multirow}   
\usepackage{multicol}   
\usepackage{caption}    
\usepackage{array}      
\usepackage{adjustbox}  
\usepackage{hyperref}
\usepackage{url}
\usepackage{pifont}
\usepackage{amsmath}    

\usepackage{amssymb}    
\usepackage{amsfonts}   
\usepackage{bm}
\usepackage{amsthm} 
\usepackage{graphicx}
\usepackage{subcaption}  
\usepackage{wrapfig}
\usepackage{svg}
\usepackage{xcolor,soul}

\svgpath{{ figures/}}         

\theoremstyle{plain} 

\usepackage{enumitem}
\newcommand{\indicator}[1]{\mathbf{1}\!\left[#1\right]}

\copyrightyear{2026}
\acmYear{2026}
\setcopyright{cc}
\setcctype{by}
\acmConference[WWW '26]{Proceedings of the ACM Web Conference 2026}{April 13--17, 2026}{Dubai, United Arab Emirates}
\acmBooktitle{Proceedings of the ACM Web Conference 2026 (WWW '26), April 13--17, 2026, Dubai, United Arab Emirates}
\acmPrice{}
\acmDOI{10.1145/3774904.3793044}
\acmISBN{979-8-4007-2307-0/2026/04}

\begin{document}

\title{Spiking Graph Predictive Coding for Reliable OOD Generalization }

\author{Jing Ren}
\authornote{Both authors contributed equally to this research.}
\orcid{0000-0003-0169-1491}
\affiliation{%
 \institution{RMIT University}
 \city{Melbourne}
 \country{Australia}
}
\email{jing.ren@ieee.org}

\author{Jiapeng Du}
\authornotemark[1]
\affiliation{%
 \institution{RMIT University}
 \city{Melbourne}
 \country{Australia}
}
\email{s4141788@student.rmit.edu.au}

\author{Bowen Li}
\orcid{0009-0007-6470-5607}
\affiliation{%
 \institution{RMIT University}
 \city{Melbourne}
 \country{Australia}
}
\email{s3890442@student.rmit.edu.au}

\author{Ziqi Xu}
\orcid{0000-0003-1748-5801}
\affiliation{%
 \institution{RMIT University}
 \city{Melbourne}
 \country{Australia}
}
\email{ziqi.xu@rmit.edu.au}

\author{Xin Zheng}
\affiliation{%
  \institution{RMIT University}
  \city{Melbourne}
  \country{Australia}}
  \email{xin.zheng2@rmit.edu.au}

\author{Hong Jia}
\affiliation{%
  \institution{University of Auckland}
  \city{Auckland}
  \country{New Zealand}}
\email{hong.jia@auckland.ac.nz}

\author{Suyu Ma}
\affiliation{%
  \institution{CSIRO's Data61}
  \city{Melbourne}
  \country{Australia}}
\email{suyu.ma@data61.csiro.au}

\author{Xiwei Xu}
\affiliation{%
  \institution{CSIRO's Data61}
  \city{Sydney}
  \country{Australia}}
\email{xiwei.xu@data61.csiro.au}

\author{Feng Xia}
\authornote{Corresponding author.}
\affiliation{%
  \institution{RMIT University}
  \city{Melbourne}
  \country{Australia}}
\email{f.xia@ieee.org}

\renewcommand{\shortauthors}{Jing Ren et al.}

\begin{abstract}
Graphs provide a powerful basis for modeling Web-based relational data, with expressive GNNs to support the effective learning in dynamic web environments.
However, real-world deployment is hindered by pervasive out-of-distribution (OOD) shifts, where evolving user activity and changing content semantics alter feature distributions and labeling criteria. These shifts often lead to unstable or overconfident predictions, undermining the trustworthiness required for Web4Good applications. Achieving reliable OOD generalization demands principled and interpretable uncertainty estimation; however, existing methods are largely post-hoc, insensitive to distribution shifts, and unable to explain where uncertainty arises especially in high-stakes settings.
To address these limitations, we introduce \underline{\textbf{S}}p\underline{\textbf{I}}king \underline{\textbf{G}}rap\underline{\textbf{H}} predic\underline{\textbf{T}}ive coding (\textbf{SIGHT}), an uncertainty-aware plug-in graph learning module for reliable OOD Generalization. SIGHT performs iterative, error-driven correction over spiking graph states, enabling models to expose internal mismatch signals that reveal where predictions become unreliable.
Across multiple graph benchmarks and diverse OOD scenarios, SIGHT consistently enhances predictive accuracy, uncertainty estimation, and interpretability when integrated with GNNs.
\end{abstract}




\maketitle

\section{Introduction}
Graphs provide a powerful abstraction for modeling relational dependencies across 
various Web-related structural data, such as social networks~\cite{ren2023graph}, recommendation platforms~\cite{xu2025towards,liu2025test}, and knowledge graphs~\cite{yu2023web,liu2020web}. To effectively model such complex and diverse graphs, graph neural networks (GNNs) have become powerful tools for mining social knowledge, identifying harmful behaviors, and enabling automated decision-making in Web environments~\cite{xia2026graph,xia2021graph,liu2022graph,khoshraftar2024survey,du2025telling}. 
Despite expressive success, graph out-of-distribution (OOD) becomes a major obstacle for deploying GNNs in practice, when test graphs deviate from the training distribution~\cite{liu2023flood,yuan2025structure}.
For example, in online social networks,
users' activity may shift suddenly from routine conversations to emergency reporting or information seeking during extreme weather incidents. These forms of OOD shift make GNNs produce unstable, inaccurate, or overconfident predictions. 
\textbf{Relevance to Web4Good.} This research problem arises within the scope of responsible AI, aiming to build well-generalized GNNs with robustness, reliability, and transparency, leading to essential benefits for deploying graph learning systems in socially critical applications, such as risk detection, content governance, and public-safety interventions~\cite{kaur2022trustworthy,liang2022advances,wang2021confident}.

Reliable graph OOD generalization aims to ensure that models can adapt to shifted distributions while producing well-calibrated uncertainty estimates that reflect prediction reliability.
Existing approaches of model uncertainty estimation for GNNs mainly face two critical challenges: \textbf{C1: Limited capability under OOD shifts}, where most methods are post-hoc and decoupled from the inference process~\cite{wu2024graph,wang2025gold}, so that the estimated uncertainty cannot explicitly adapt to distributional shifts; \textbf{C2: Limited interpretability of uncertainty sources}, where most methods lacks insight into where uncertainty arises, making it hard to determine when predictions should be trusted or overridden~\cite{li2025tackling,trivedi2024accurate}. Recent progress in incorporating the neuroscience concept of predictive coding (PC)~\citep{rao1999predictive} into neural network design has demonstrated improved model calibration performance~\citep{byiringiro2022robust}. However, directly integrating PC with traditional GNNs is challenging under OOD scenarios, as dense synchronous updates conflict with PC’s iterative, event-driven dynamics.
Hence, addressing these challenges is essential for building trustworthy and socially aligned graph learning systems in dynamic web environments~\cite{han2025uncertainty,chen2025uncertainty}.

In light of this, we propose a novel framework of \underline{\textbf{S}}p\underline{\textbf{I}}king \underline{\textbf{G}}rap\underline{\textbf{H}} predic\underline{\textbf{T}}ive coding, dubbed \textbf{SIGHT}, which enables predictive coding on graphs to operate through sparse and asynchronous spiking dynamics that naturally support local error correction and biologically inspired inference.
Specifically, our proposed SIGHT contains two essential components: (1) \textit{Predictive coding module}, which embeds predictive coding dynamics directly into the inference process, enabling uncertainty to emerge from the model’s internal error-correction trajectory (addressing C1). (2) \textit{Spiking error-propagation mechanism}, which exposes interpretable internal signals at each inference step. These sparse and asynchronous spikes explicitly reveal where and how the model struggles to reconcile conflicting information across the graph (addressing C2). 
SIGHT serves as an uncertainty-aware plug-in module for graph learning, designed to support reliable OOD generalization while maintaining computational efficiency and intrinsic interpretability.

In summary, the contributions of this work are presented below:
\begin{itemize}[leftmargin=1.2em, itemsep=3pt, topsep=3pt]
    \item \textbf{Practical Research Problem.} We focus on the \textit{GNN model uncertainty estimation} problem under the practical graph OOD scenarios, contributing to developing more robust, reliable, and interpretable predictions.
    \item \textbf{Novel Framework.} We first develop \textbf{SIGHT}, a lightweight plug-in GNN uncertainty estimation model, which integrates spiking neural dynamics with predictive coding to perform iterative, error-driven corrections on graph predictions, offering transparent interpretability for uncertainty sources with accountable decision-making.
    \item \textbf{Extensive Experiments.} We conduct experiments across multiple benchmarks and GNN backbones, showing that our proposed SIGHT achieves superior OOD generalization ability and more accurate uncertainty estimation performance.
\end{itemize}

\section{Related work}
 \textbf{Graph OOD Generalization and Uncertainty Estimation.} OOD Generalization and uncertainty estimation have become increasingly important in graph learning, especially as high-stakes social systems rely on AI models to analyze relational data collected from dynamic and often unstable environments~\cite{wu2022handling,li2025out}. In real-world web settings, distribution shifts frequently arise due to changes in node attributes, evolving interaction patterns (i.e., covariate shift), or shifting societal contexts that alter the relationship between features and labels (i.e., concept shift)~\cite{bazhenov2023evaluating,liu2023structural,liu2023flood,li2025out}. These shifts can substantially degrade model reliability, leading to misleading predictions or masking critical uncertainty in decisions that affect public welfare and vulnerable communities. Conventional GNNs remain particularly susceptible to these challenges, as they are typically trained to fit the observed distribution and lack principled mechanisms to detect, express, or reason about uncertainty under shift~\cite{yehudai2021local,wu2022handling,yang2022learning,yu2023mind,yuan2025structure}. Consequently, their predictions are often overconfident and provide limited insight into when and why the model struggles, hindering meaningful interpretability and reducing trustworthiness in web applications.

\noindent
\textbf{Uncertainty Estimation.} Existing approaches largely fall into two categories: post-hoc calibration~\cite{guo2017calibration,kull2019beyond,wang2021confident,hsu2022makes}, which adjusts confidence after training, and intrinsic estimation techniques~\cite{trivedi2024accurate,thiagarajan2022single}, which modify training to yield better-calibrated predictions. Post-hoc calibration models, such as temperature scaling~\cite{guo2017calibration}, Bayesian graph models~\cite{gal2016dropout}, and deep ensembles~\cite{lakshminarayanan2017simple}, improve the alignment between predicted probabilities and empirical correctness by adjusting model confidence after training, but they often fail under distribution shifts. By contrast, \cite{trivedi2024accurate} propose an intrinsic method, G-$\Delta$UQ, which modifies the training process to produce calibrated predictions. However, these approaches offer limited explanation of where uncertainty originates, despite the fact that such interpretability is critical for trustworthy deployment in high-stake settings. 

\noindent\textbf{Predictive Coding.}
PC is a neurobiological theory proposing that the brain continually generates top-down predictions of sensory input, while bottom-up signals convey only the prediction errors, thereby driving perception and learning~\cite{hebb1949organization, rao1999predictive}. This biologically inspired alternative to backpropagation (BP)~\cite{friston2018does,salvatori2023survey} has been applied in generative modeling~\cite{ororbia2022neural}, continual learning~\cite{ororbia2020continual}, reinforcement learning~\cite{ororbia2023active}, graph learning~\cite{byiringiro2022robust}, and arbitrary network learning~\cite{salvatori2022learning}. Moreover,~\cite{byiringiro2022robust} find that replacing traditional BP in GNNs with PC can achieve better calibration than traditional GNNs in ID scenarios. However, predictive coding has not yet been explored in shifted graph environments where generalization ability, reliable decision-making, and uncertainty transparency are essential for deployment in high-stakes social systems.


\section{Methodology} 
\label{sec:method}
\begin{figure*}[t]
    \centering

        \centering
        \includegraphics[width=\linewidth]{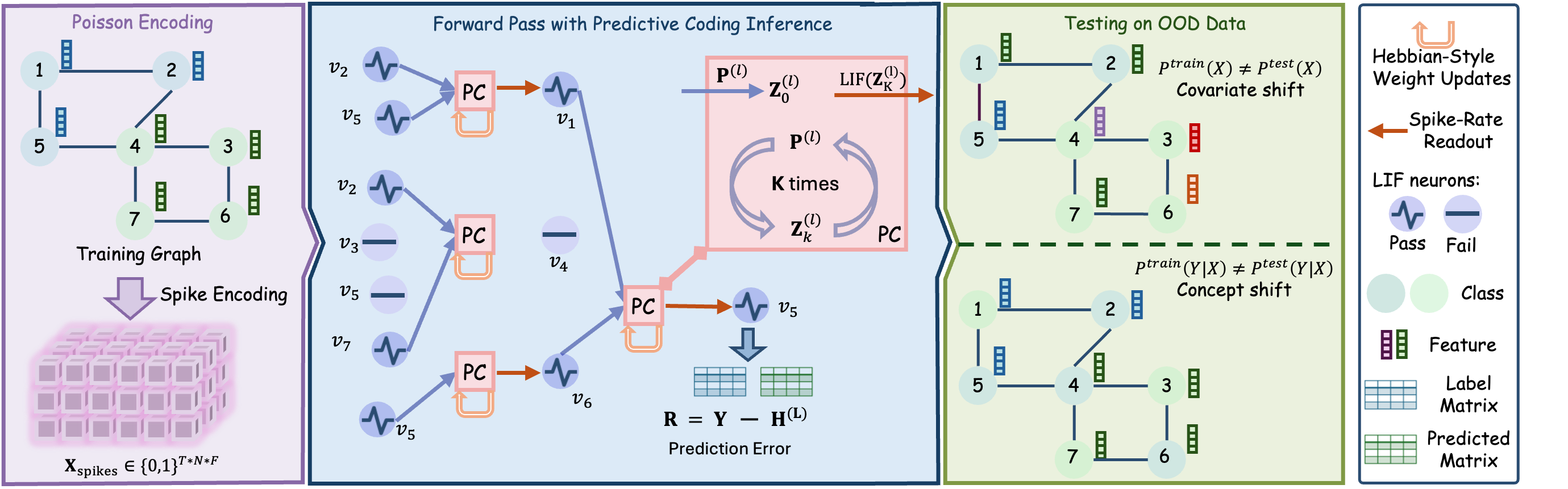} 
        \label{fig:sub1}

    \caption{ Overview of the proposed framework SIGHT. In training, the encoded node representations are learned by aggregating neighbor information through GNNs. Before propagation, predictions are refined via an iterative predictive coding (PC) unit over LIF neurons. In testing, the model will be tested on datasets with \textit{covariate shift}, where the input feature distribution changes while the underlying labels remain stable (i.e., $P^{\text{train}}(X)\neq P^{\text{test}}(X)$ but $P^{\text{train}}(Y|X)=P^{\text{test}}(Y|X)$), and with \textit{concept shift}, where the meaning of labels or the feature–label relationship itself changes over time (i.e., $P^{\text{train}}(Y|X)\neq P^{\text{test}}(Y|X)$).}
    \label{fig:framework}
\end{figure*}

In this paper, we propose a flexible framework for spiking graph predictive coding, dubbed \textbf{SIGHT}, for reliable uncertainty estimation in GNN-based classification. Figure~\ref{fig:framework} illustrates the overall framework of SIGHT. 
Specifically, SIGHT performs uncertainty-aware graph learning by integrating spiking dynamics with predictive coding. We first encode continuous node features into biologically plausible spike trains using Poisson encoding, and then conduct a forward pass through iterative predictive coding inference, where node states are refined via local mismatch correction. The resulting spike trains are aggregated through spike-rate readout to obtain stable representations, and model parameters are updated using local Hebbian-style rules driven by prediction errors, enabling efficient, interpretable, and OOD-sensitive learning

\subsection{Preliminary}
\noindent{\textbf{Notations}.}
We consider a graph $\mathcal{G}=(\mathcal{V},\mathcal{E})$ composed of a node set $\mathcal{V}$ and an edge set $\mathcal{E}$, where $|\mathcal{V}|=N$ and the graph is represented by an adjacency matrix $\textbf{A} \in \mathbb{R}^{N\times N}$. Each node $v_i \in \mathcal{V}$ is associated with a feature vector $\mathbf{x}_i \in \mathbb{R}^F$, which together form the feature matrix $\textbf{X} \in \mathbb{R}^{N\times F}$. 
We adopt the symmetrically normalised adjacency matrix $\tilde{\textbf{A}} = \textbf{D}^{-\tfrac{1}{2}}(\textbf{A} + \textbf{I})\textbf{D}^{-\tfrac{1}{2}}$, where $\textbf{D}$ is the degree matrix. An $L$-layer SIGHT model maintains hidden dimensions $d_0=F, d_1, \dots, d_L$ and weight matrices $\textbf{W}^{(l)} \in \mathbb{R}^{d_{l-1}\times d_l}$. We focus on node classification, where each node $v_i$ is assigned a label $y_i \in {1,\dots,C}$, and the labels of all nodes are collectively represented by a one-hot matrix $\mathbf{Y} \in {0,1}^{N\times C}$. 

\noindent\textbf{Problem Definition.} Given an input graph $\mathcal{G}=(\mathcal{V},\mathcal{E})$ with out-of-distribution problem $\mathcal{P}_{\text{test}} \neq \mathcal{P}_{\text{train}}$,
the problem of GNN model uncertainty estimation can be defined by learning a projection function for node $v$ as
\begin{equation}
f_{\boldsymbol{\theta}} : (\mathcal{G}, v) \rightarrow (\hat{\textbf{y}}_v, u_v),
\end{equation}
where \( \hat{\textbf{y}}_v \) is the predicted class label for node \(v\), and \( u_v \) is a confidence score that reflects the reliability of the prediction under distribution shift.


\subsection{Poisson Encoding of Node Features}

To support spiking-based inference, we use Poisson encoding for node feature initialization, so that continuous features are transformed into biologically plausible spike trains. Given the node feature matrix $\mathbf{X}\in\mathbb{R}^{N\times F}$, 
we first normalise each feature channel to the range $[0,1]$, yielding $\tilde{\mathbf{X}}$. 
Each scalar feature $\tilde{X}_{i,f}$ is then converted into a spike train using a Poisson process:
\begin{equation}\label{eq:spike}
  \mathbf{X}_{\mathrm{spikes}}[t,i,f] \sim \mathrm{Bernoulli}\big(\tilde{\mathbf{X}}_{i,f}\big), 
  \quad \mathbf{X}_{\mathrm{spikes}} \in \{0,1\}^{T\times N\times F}.
\end{equation}
At each simulation step $t$, the slice $\mathbf{X}_{\mathrm{spikes}}[t]$ is used as the network input.

\subsection{Forward Pass with Predictive Coding Inference}
Without loss of generality, we use GCN~\cite{kipf2017semi} as the backbone in our SIGHT, and it can be readily extended to other GNN architectures, such as GAT~\cite{velivckovic2018graph}.
The forward pass integrates graph convolution with predictive coding dynamics. 
For each layer $l \in \{1,\dots,L\}$ and timestep $t$, the procedure consists of:

\paragraph{Graph Convolution.}
We compute latent predictions as:
\begin{equation}\label{eq:convolution}
    \mathbf{P}^{(l)} = \tilde{\mathbf{A}} \mathbf{H}^{(l-1)} \mathbf{W}^{(l)},
    \quad \mathbf{H}^{(0)} = \mathbf{X}_{\mathrm{spikes}}[t].
\end{equation}

\paragraph{Predictive Coding Loop.}
We initialise the latent state $\mathbf{Z}^{(l)}_0 \leftarrow \mathbf{P}^{(l)}$ 
and perform $K$ inference iterations:
\begin{equation}\label{eq:pc}
    \mathbf{E}_k^{(l)} = \mathbf{P}^{(l)}-\mathbf{Z}_k^{(l)}  ,  ~~~
   \mathbf{U}_k = \mathbf{P}^{(l)} + \gamma \, \mathrm{LIF}_{\mathrm{err}}(\mathbf{E}_k^{(l)}), ~~~\mathbf{Z}_{k+1}^{(l)} = \mathrm{LIF}_{\mathrm{pred}}(\mathbf{U}_k).
\end{equation}
   
Here, $\textbf{Z}^{(l)} \in \mathbb{R}^{N\times d_l}$ denotes layerwise latent predictions, $\textbf{P}^{(l)} \in \mathbb{R}^{N\times d_l}$ denotes instantaneous pre-activations, and $\mathbf{E}_k^{(l)}$ denotes the local residual, $\gamma$ is a correction gain, and $\mathrm{LIF}_{\mathrm{err}}, \mathrm{LIF}_{\mathrm{pred}}$ denote leaky integrate-and-fire dynamics. 

After $K$ iterations, the final latent state $\mathbf{Z}_K^{(l)}$ serves as the layer output. The spike tensor $\mathbf{Z}_K^{(l)}$ is rectified and propagated to the next layer. At the final layer, the corrected spikes are passed to the classifier. This iterative procedure can also be interpreted from an energy-minimisation perspective, where the dynamics of LIF neurons approximate gradient descent on a local error functional.

\paragraph{Uncertainty Quantification via Error Statistics.}
We define the predictive coding confidence at layer $l$ for node $i$ as $u_i^{(l)} = \exp\big(-|\mathbf{E}^{(l)}_{i,:}|_2^2\big),$ which approximates the likelihood of observation under Gaussian residual noise. This provides a principled mapping from error magnitudes to epistemic uncertainty, eliminating the need for explicit ensembling or Monte Carlo sampling.
\begin{table*}[t]
\centering
\caption{Node classification accuracy and uncertainty on ID and OOD datasets with the GCN backbone. Best results are in \textbf{bold}, and SIGHT is marked with \colorbox{gray!20}{\hspace{1em}\vphantom{a}}. Results with the GAT backbone are in Table~\ref{tab:PerformanceGAT}.}
\label{tab:PerformanceGCN}
\setlength{\tabcolsep}{3pt}
\resizebox{\linewidth}{!}{
\begin{tabular}{c l
                cc cc  
                cc cc cc} 
\toprule
\multicolumn{1}{c}{\multirow{2}{*}{}} & \multicolumn{1}{c}{\multirow{2}{*}{\textbf{Method}}}
& \multicolumn{2}{c}{\textbf{Accuracy} $\uparrow$}
& \multicolumn{2}{c}{\textbf{ECE} $\downarrow$}
& \multicolumn{2}{c}{\textbf{NLL} $\downarrow$}
& \multicolumn{2}{c}{\textbf{BS} $\downarrow$}
& \multicolumn{2}{c}{\textbf{AUROC} $\uparrow$}
\\\cmidrule(lr){3-4}\cmidrule(lr){5-6}\cmidrule(lr){7-8}\cmidrule(lr){9-10}\cmidrule(lr){11-12}
& & ID & OOD & ID & OOD & ID & OOD & ID & OOD & ID & OOD \\
\midrule

\multirow{3}{*}{\rotatebox{90}{\textbf{Cora}}}
 & GCN        &89.8$\pm$0.3 &43.5$\pm$5.2 &0.019$\pm$0.003 &0.391$\pm$0.071 &0.318$\pm$0.008 &2.595$\pm$0.512 &0.150$\pm$0.004 &0.919$\pm$0.103 &\textbf{88.7$\pm$0.3} &59.9$\pm$3.7\\
 & ~+G-$\Delta$UQ       &91.3$\pm$0.4 &80.1$\pm$4.7 &0.019$\pm$0.004 &0.053$\pm$0.031 &0.270$\pm$0.018 &0.574$\pm$0.149 &0.130$\pm$0.007 &0.288$\pm$0.068 &87.3$\pm$1.2 &81.5$\pm$2.8 \\  
 & \cellcolor{gray!20}~+SIGHT           &\cellcolor{gray!20}\textbf{93.2$\pm$0.6}  &\cellcolor{gray!20}\textbf{95.6$\pm$0.2}  &\cellcolor{gray!20}\textbf{0.015$\pm$0.002}  &\cellcolor{gray!20}\textbf{0.009$\pm$0.002}  &\cellcolor{gray!20}\textbf{0.241$\pm$0.021}  &\cellcolor{gray!20}\textbf{0.153$\pm$0.010}  &\cellcolor{gray!20}\textbf{0.105$\pm$0.009}  &\cellcolor{gray!20}\textbf{0.068$\pm$0.003}  &\cellcolor{gray!20}86.7$\pm$1.2 &\cellcolor{gray!20}\textbf{89.8$\pm$1.5}\\
\midrule

\multirow{3}{*}{\rotatebox{90}{\textbf{Citeseer}}}
 & GCN        &82.1$\pm$0.4  &46.6$\pm$1.9  &0.026$\pm$0.003  &0.334$\pm$0.027  &0.503$\pm$0.012 &1.838$\pm$0.051  &0.252$\pm$0.006 &0.793$\pm$0.026&\textbf{84.1$\pm$0.8}&74.1$\pm$1.6  \\
 & ~+G-$\Delta$UQ           &81.6$\pm$0.4  &71.0$\pm$4.2  &\textbf{0.021$\pm$0.008}  &0.066$\pm$0.027  &0.531$\pm$0.037  &0.889$\pm$0.145 &0.263$\pm$0.010 &0.416$\pm$0.056 &82.2$\pm$1.6 &76.5$\pm$1.3\\  
 & \cellcolor{gray!20}~+SIGHT         &\cellcolor{gray!20}\textbf{86.6$\pm$0.7}  &\cellcolor{gray!20}\textbf{91.2$\pm$0.4}  &\cellcolor{gray!20}0.028$\pm$0.006  &\cellcolor{gray!20}\textbf{0.024$\pm$0.004} & \cellcolor{gray!20}\textbf{0.452$\pm$0.023}  &\cellcolor{gray!20}\textbf{0.307$\pm$0.016}  &\cellcolor{gray!20}\textbf{0.204$\pm$0.010}  &\cellcolor{gray!20}\textbf{0.136$\pm$0.008} & \cellcolor{gray!20}80.1$\pm$1.1 &\cellcolor{gray!20}\textbf{83.4$\pm$0.8}\\
\midrule

\multirow{3}{*}{\rotatebox{90}{\textbf{Pubmed}}}

 & GCN        & 88.0$\pm$0.1  &73.7$\pm$1.9  &0.008$\pm$0.001  &0.146$\pm$0.028  &0.306$\pm$0.001&0.960$\pm$0.081  &0.173$\pm$0.000 &0.411$\pm$0.028&85.2$\pm$0.3&72.3$\pm$1.5  \\
 & ~+G-$\Delta$UQ           &91.3$\pm$0.2  &83.8$\pm$0.7  &\textbf{0.006$\pm$0.001}  &0.090$\pm$0.009 &\textbf{0.236$\pm$0.002} &0.668$\pm$0.048 &\textbf{0.129$\pm$0.002} &0.263$\pm$0.013 &\textbf{86.3$\pm$0.4} &74.2$\pm$0.3\\  
 & \cellcolor{gray!20}~+SIGHT           &\cellcolor{gray!20}\textbf{90.0$\pm$0.1}  &\cellcolor{gray!20}\textbf{90.1$\pm$0.2}  &\cellcolor{gray!20}0.009$\pm$0.001  &\cellcolor{gray!20}\textbf{0.006$\pm$0.001}  &\cellcolor{gray!20}0.275$\pm$0.001  &\cellcolor{gray!20}\textbf{0.275$\pm$0.004}  &\cellcolor{gray!20}0.149$\pm$0.001  &\cellcolor{gray!20}\textbf{0.148$\pm$0.002}  &\cellcolor{gray!20}85.5$\pm$0.4 &\cellcolor{gray!20}\textbf{85.1$\pm$0.2}\\
\midrule

\multirow{3}{*}{\rotatebox{90}{\textbf{Twitch}}}

 & GCN        &60.7$\pm$9.0  &57.5$\pm$4.1  & 0.129$\pm$0.066  &0.046$\pm$0.028  &0.685$\pm$0.006&0.683$\pm$0.007  & 0.492$\pm$0.006 &0.490$\pm$0.008&\textbf{59.5$\pm$4.3}& 49.2$\pm$3.3 \\
 & ~+G-$\Delta$UQ           &\textbf{64.8$\pm$9.9}  &58.3$\pm$2.9  &0.163$\pm$0.067  &0.065$\pm$0.024 &0.690$\pm$0.021  &0.687$\pm$0.009& 0.497$\pm$0.021 &0.494$\pm$0.009 &58.0$\pm$9.1 &45.4$\pm$1.5 \\  
 & \cellcolor{gray!20}~+SIGHT         & \cellcolor{gray!20}59.3$\pm$6.6  &\cellcolor{gray!20}\textbf{60.6$\pm$1.4}  &\cellcolor{gray!20}\textbf{0.077$\pm$0.033}  &\cellcolor{gray!20}\textbf{0.039$\pm$0.009}  &\cellcolor{gray!20}\textbf{0.676$\pm$0.035}  &\cellcolor{gray!20}\textbf{0.671$\pm$0.005}  &\cellcolor{gray!20}\textbf{0.483$\pm$0.033}  &\cellcolor{gray!20}\textbf{0.478$\pm$0.005}  &\cellcolor{gray!20}56.1$\pm$6.9 &\cellcolor{gray!20}\textbf{52.2$\pm$3.9}\\
\midrule

\multirow{3}{*}{\rotatebox{90}{\textbf{CBAS}}}

 & GCN        & 77.0$\pm$4.9 & 69.6$\pm$4.9 &0.095$\pm$0.025  & 0.116$\pm$0.007 &0.596$\pm$0.032& 0.840$\pm$0.026 &0.327$\pm$0.025 &0.432$\pm$0.022&77.5$\pm$5.0& 71.6$\pm$2.6\\
 & ~+G-$\Delta$UQ           &76.4$\pm$5.2&65.1$\pm$2.8  &0.125$\pm$0.018 &0.131$\pm$0.039&0.611$\pm$0.058 &0.881$\pm$0.043   &0.337$\pm$0.036&0.461$\pm$0.031 &77.4$\pm$6.4 &66.9$\pm$7.4\\     
 & \cellcolor{gray!20}~+SIGHT          &\cellcolor{gray!20}\textbf{93.9$\pm$2.2} &\cellcolor{gray!20}\textbf{76.4$\pm$2.8}  &\cellcolor{gray!20}\textbf{0.071$\pm$0.021}  &\cellcolor{gray!20}\textbf{0.086$\pm$0.020}  &\cellcolor{gray!20}\textbf{0.241$\pm$0.067}&\cellcolor{gray!20}\textbf{0.746$\pm$0.075}  & \cellcolor{gray!20}\textbf{0.114$\pm$0.034}  &\cellcolor{gray!20}\textbf{0.363$\pm$0.032} &\cellcolor{gray!20}\textbf{81.6$\pm$4.9}&\cellcolor{gray!20}\textbf{73.0$\pm$1.8}\\
\bottomrule
\end{tabular}
}
\end{table*}

\subsection{Spike-Rate Readout}

We adopt LIF neurons due to their biological plausibility and ability to capture temporal dynamics with lightweight, event-driven computation. Since each LIF neuron~\cite{gerstner2002spiking} produces binary spikes, 
we extract rate codes by averaging across $T$ timesteps
\begin{equation}
    \mathbf{H}^{(l)} = \frac{1}{T}\sum_{t=1}^T \mathbf{Z}^{(l)}_t,
\end{equation}where $\mathbf{Z}^{(l)}_t \in \{0,1\}^{N\times d_l}$ is the spike matrix at timestep $t$. These rate codes provide stable representations for downstream prediction and learning.

\subsection{Local Hebbian-Style Weight Updates}

Local Hebbian updates allow SIGHT to perform biologically plausible, computationally efficient, and interpretable error-driven learning, without requiring global backpropagation. They align naturally with predictive coding dynamics and support event-driven spiking computation during OOD inference.
At the top layer $L$, the prediction error is defined as
\begin{equation}
    \mathbf{R}^{(L)} = \mathbf{Y} - \mathbf{H}^{(L)},
\end{equation} where $\mathbf{Y}$ is the one-hot label matrix. 

For each layer $l$, weights are updated as
$\Delta \mathbf{W}^{(l)} \propto (\mathbf{H}^{(l-1)})^\top \mathbf{R}^{(l)}.$ This rule couples presynaptic rates $\mathbf{H}^{(l-1)}$ with postsynaptic residuals $\mathbf{R}^{(l)}$, implementing a biologically plausible outer-product Hebbian update. Crucially, no backpropagated error is required and each layer learns from its own residuals.


\subsection{Computational Complexity Analysis}
We conduct energy efficiency analysis and time complexity analysis. 

\paragraph{Energy Efficiency Analysis.}

Spiking implementations introduce event-driven sparsity with only a fraction $\rho \ll 1$ of neurons fire per timestep. The effective cost per timestep is therefore $\mathcal{O}\big(\rho |\mathcal{E}| d_{l-1} + \rho N d_l\big).$
This makes SIGHT well-suited for neuromorphic hardware, where multiplications are replaced by accumulations triggered by spikes, reducing both latency and power consumption. 

\paragraph{Time Complexity.}  
For each layer $l$ with input dimension $d_{l-1}$ and output dimension $d_l$, a single graph convolution requires $\mathcal{O}\big(|\mathcal{E}|d_{l-1} + N d_l\big),$ where $|\mathcal{E}|$ is the number of edges. Predictive coding inference introduces an additional factor of $K$ inner iterations per layer. Thus, the per-layer cost is $\mathcal{O}\big(K(|\mathcal{E}|d_{l-1} + N d_l)\big).$  Over $L$ layers and $T$ Poisson timesteps, the total complexity is $\mathcal{O}\big(TK \sum_{l=1}^L (|\mathcal{E}|d_{l-1} + N d_l)\big).$ In practice, $T$ and $K$ are small constants, so the overhead remains linear in $|\mathcal{E}|$ and $N$.


\paragraph{Comparison with Backpropagation.}  
Backpropagation requires gradient chaining across all layers and timesteps, incurring $\mathcal{O}(LT)$ memory and repeated backward passes. SIGHT avoids this by relying on local Hebbian updates $ \Delta \textbf{W}^{(l)} \propto (\textbf{H}^{(l-1)})^\top \textbf{R}^{(l)},$ which only requires forward states and local residuals. This significantly reduces both runtime constants and memory footprint.

SIGHT achieves comparable asymptotic complexity to conventional GNNs while avoiding the heavy memory burden of backpropagation. Event-driven sparsity further improves efficiency, enabling scalable, energy-aware training and inference for large graphs and safety-critical applications.

\section{Experiments}
We conduct extensive experiments to evaluate SIGHT on five benchmark datasets. Our evaluation considers both node classification accuracy and the reliability of uncertainty estimation in ID and OOD scenarios. Additional experimental results are provided in Appendix~\ref{app:results}.

\subsection{Experimental Setup}
\label{subsec:exsetup}
\paragraph{Datasets.}
In this paper, the experiments are conducted on five node classification datasets, i.e., Cora, Citeseer, Pubmed, Twitch, and CBAS. We evaluate SIGHT under two types of distribution shifts: covariate shift and concept shift. For citation networks (i.e., Cora, Citeseer, and PubMed), we apply covariate shift by following a commonly used benchmark~\cite{wu2022handling}. Specifically, the original node labels retain while synthetically generating spurious node features to induce distribution shifts between the ID and OOD datasets, i.e., $P^{\text{train}}(X)\neq P^{\text{test}}(X)$ but $P^{\text{train}}(Y|X)=P^{\text{test}}(Y|X)$. In contrast, for Twitch and CBAS~\cite{gui2022good}, we apply concept shift by altering the conditional distribution while keeping the input distribution stable, i.e., $P^{\text{train}}(Y|X)\neq P^{\text{test}}(Y|X)$. 

\begin{figure}[t]
 
  \begin{subfigure}{0.27\textwidth}
    \centering
    \includegraphics[width=\linewidth]{ 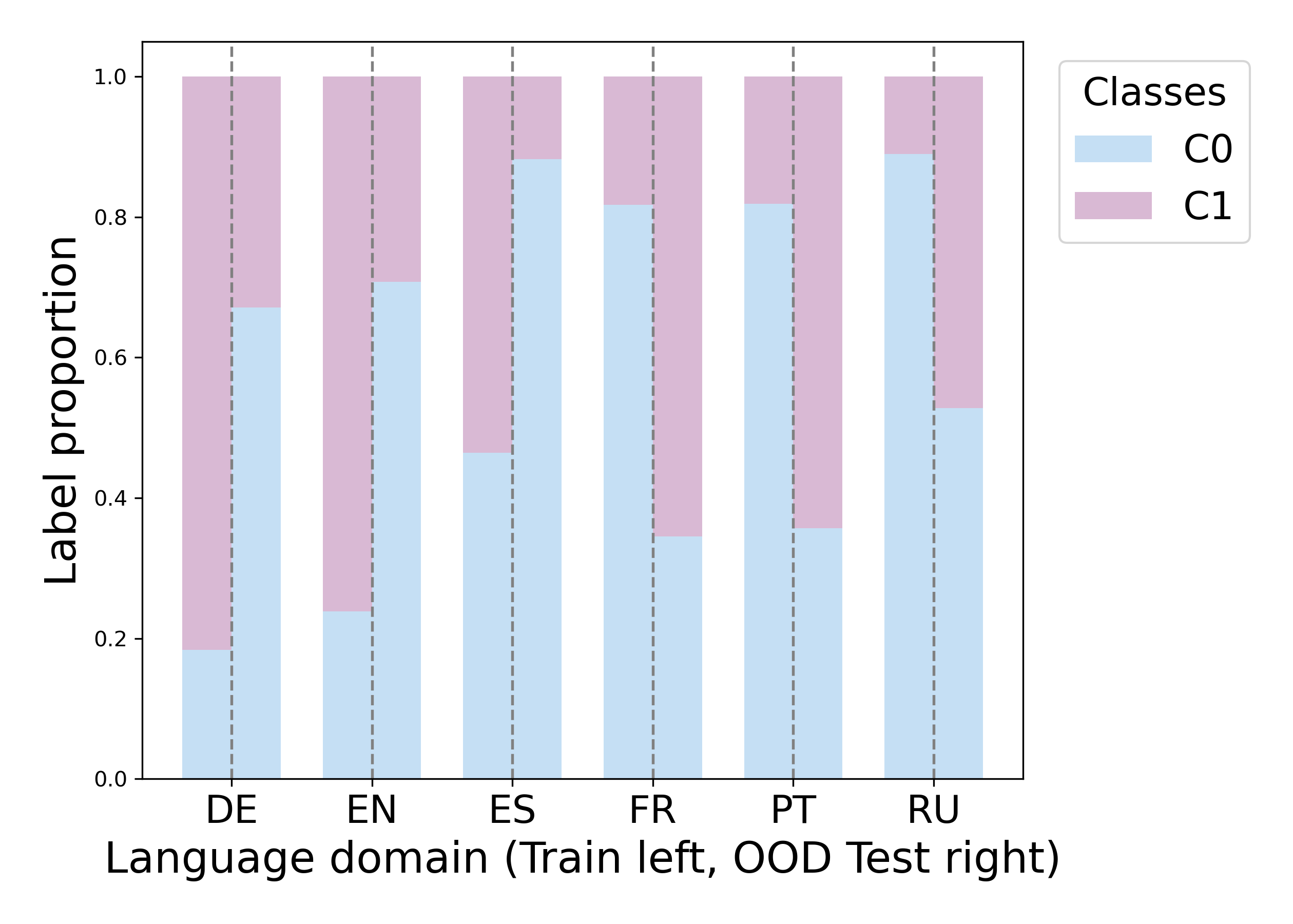} 
    \caption{Twitch}
    \label{fig:sub1}
  \end{subfigure}\hfill
  \begin{subfigure}{0.20\textwidth}
    \centering
    \includegraphics[width=\linewidth]{ 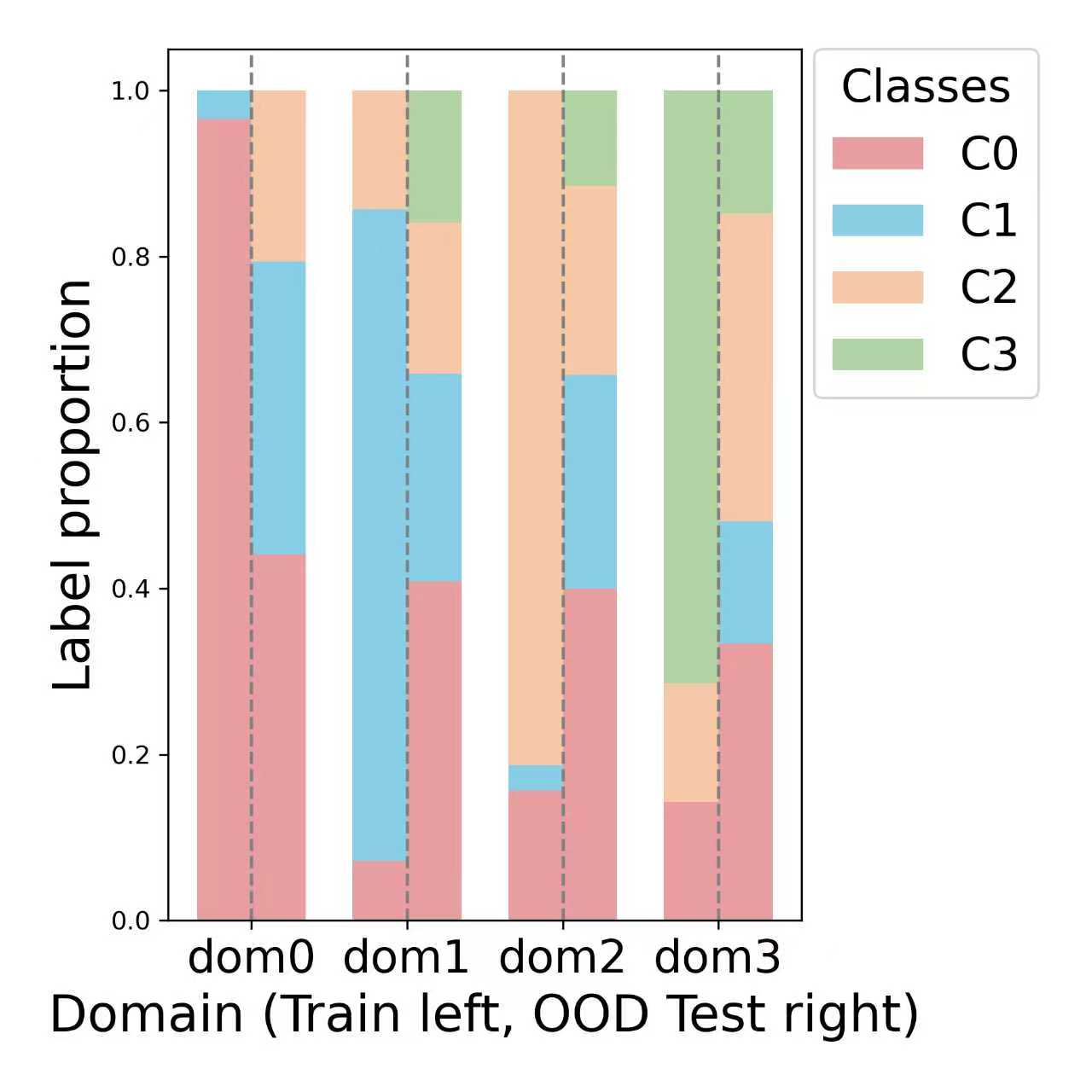}
    \caption{CBAS}
    \label{fig:sub2}
  \end{subfigure}\hfill
    \caption{Comparison of label distribution between training and testing sets on Twitch and CBAS.}
  \label{fig:conshift}
\end{figure}

\begin{figure*}[t]
  \centering
  \begin{subfigure}{0.32\textwidth}
    \centering
    \includegraphics[width=\linewidth]{ 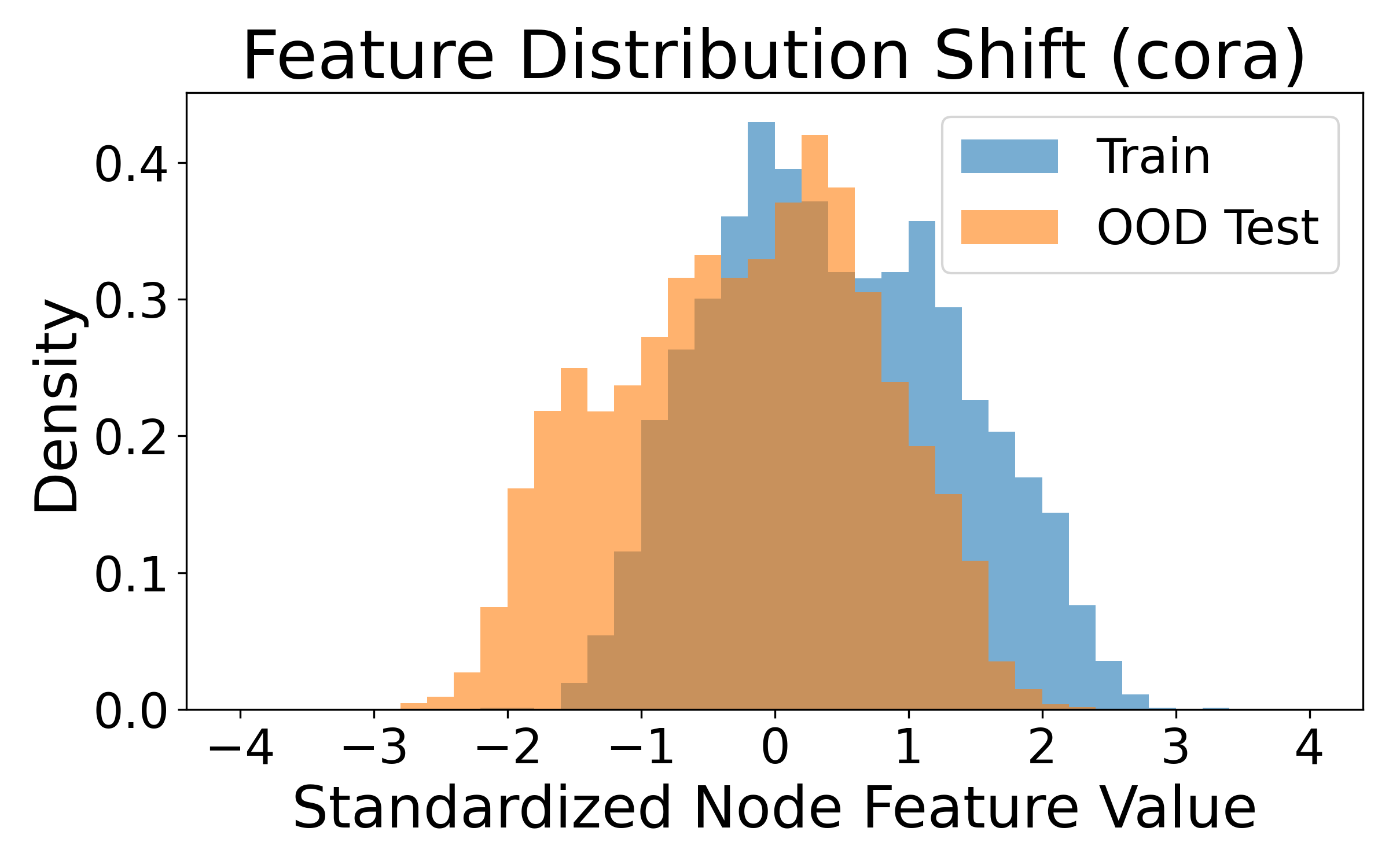} 
    \caption{Cora}
    \label{fig:sub1}
  \end{subfigure}\hfill
  \begin{subfigure}{0.32\textwidth}
    \centering
    \includegraphics[width=\linewidth]{ 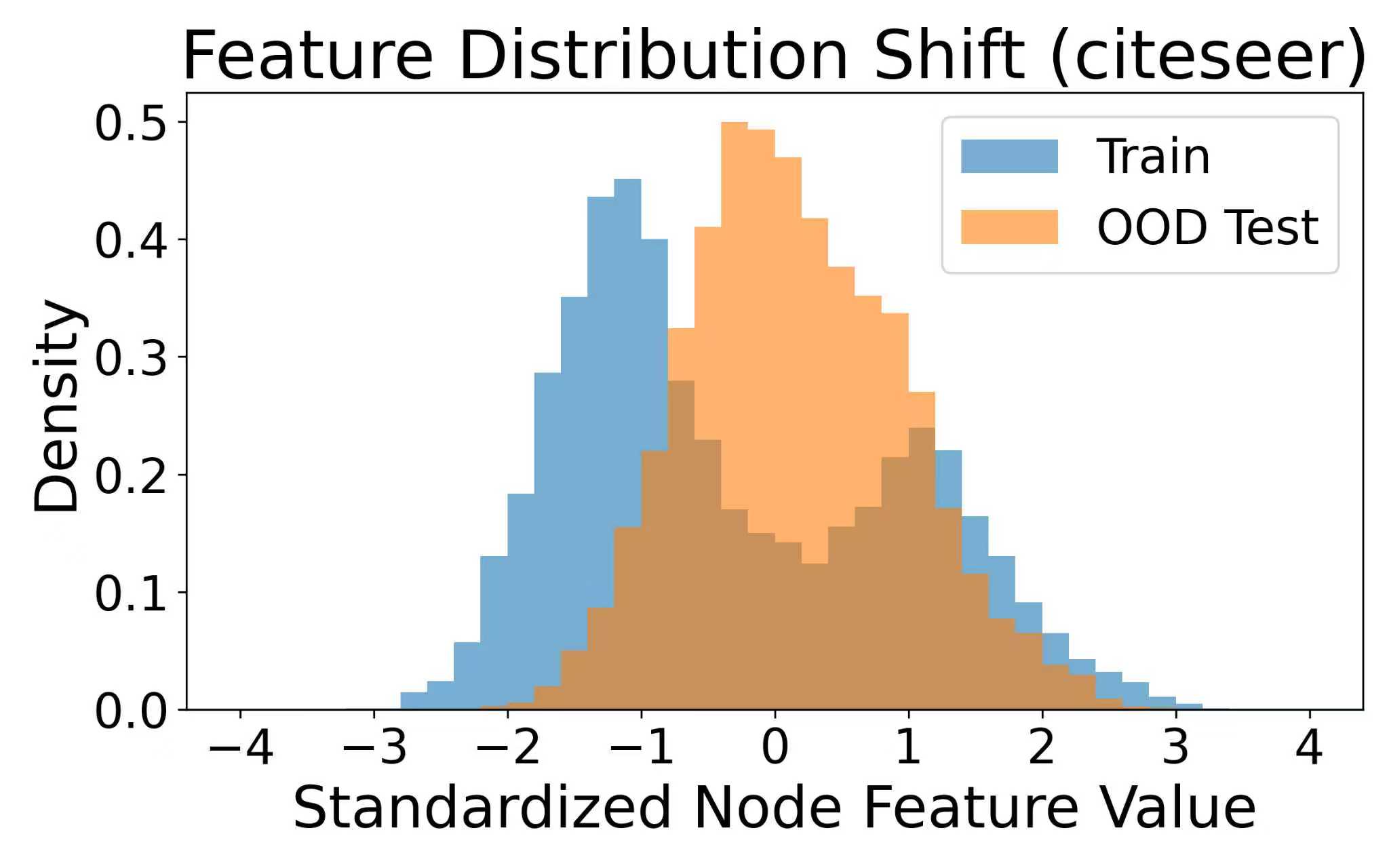}
    \caption{Citeseer}
    \label{fig:sub2}
  \end{subfigure}\hfill
  \begin{subfigure}{0.32\textwidth}
    \centering
    \includegraphics[width=\linewidth]{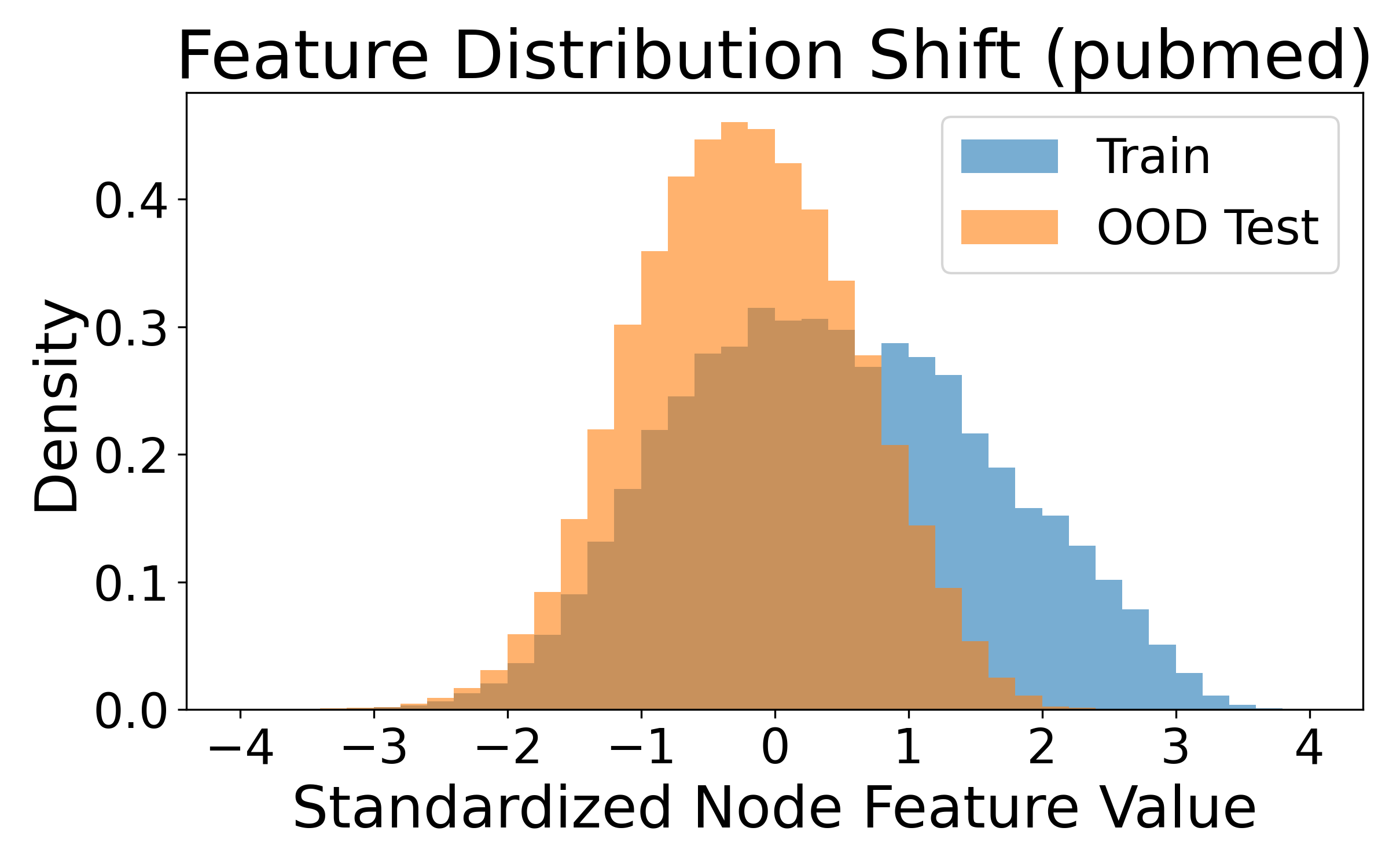}
    \caption{Pubmed}
    \label{fig:sub3}
  \end{subfigure}

    \caption{Comparison of feature distribution between training and testing sets on Cora, Citeseer, and Pubmed.}
  \label{fig:covshift}
\end{figure*}

For all datasets, we follow the dataset splits in \cite{wu2024graph} and \cite{gui2022good} for training, validation, ID and OOD testing. 
More details of the dataset statistics are shown in Table~\ref{tab:dataset}, and the label distribution between training and testing sets are shown in Figure~\ref{fig:covshift} for covariate shift of Cora, Citeseer, and Pubmed, and Figure~\ref{fig:conshift} for concept shift of Twitch and CBAS.

\begin{table}
    \centering
    \caption{Statistics of the experimental datasets.}
    \small
    \begin{tabular}{cccccc}
    \toprule
         Dataset& \# Nodes & \# Edges & \# Classes &  \# Features & Shift Type \\
         \midrule
         Cora     & 2,708  & 5,429  & 7 & 1,433 &Covariate
          \\
    Citeseer & 3,327  & 4,732  & 6 & 3,703 &Covariate
        \\
    Pubmed   & 19,717 & 44,338 & 3 &   500 &Covariate
          \\
         Twitch& 34120 & 892346& 2 & 128  &Concept \\
         CBAS& 700   & 3962  & 4 & 4 & Concept  \\
         \bottomrule
    \end{tabular}
    \label{tab:dataset}
\end{table}

\begin{table*}[t]
\centering
\small
\caption{Accuracy and ECE under distribution shifts for different post-hoc calibration models using GCN and GAT backbones, w/o and w/ SIGHT (\colorbox{gray!20}{\hspace{1em}\vphantom{a}}). \ding{55} denotes no post-hoc calibration.}
\label{tab:calibration}
\setlength{\tabcolsep}{5pt}
\renewcommand{\arraystretch}{0.75}
\begin{tabular}{l l
                cc cc  
                cc cc }
\toprule
& & \multicolumn{4}{c}{\textbf{GCN}} & \multicolumn{4}{c}{\textbf{GAT}} \\
\cmidrule(lr){3-6}\cmidrule(lr){7-10}
\multicolumn{1}{c}{\multirow{2}{*}{}} & \multicolumn{1}{l}{\multirow{2}{*}{\textbf{Model}}}
& \multicolumn{2}{c}{\textbf{Accuracy} $\uparrow$}
& \multicolumn{2}{c}{\textbf{ECE} $\downarrow$}
& \multicolumn{2}{c}{\textbf{Accuracy} $\uparrow$}
& \multicolumn{2}{c}{\textbf{ECE} $\downarrow$} \\ \cmidrule(lr){3-4} \cmidrule(lr){5-6} \cmidrule(lr){7-8}\cmidrule(lr){9-10}
& & w/o SIGHT & w/ SIGHT & w/o SIGHT & w/ SIGHT & w/o SIGHT & w/ SIGHT & w/o SIGHT & w/ SIGHT\\
\midrule

\multirow{8}{*}{\rotatebox{90}{\textbf{Cora}}}
& \ding{55}& 43.5$\pm$5.2 &\cellcolor{gray!20}95.6$\pm$0.2& 0.391$\pm$0.071 &\cellcolor{gray!20}0.009$\pm$0.002& 67.4$\pm$4.1&\cellcolor{gray!20}96.3$\pm$0.7 &  0.164$\pm$0.026&\cellcolor{gray!20}0.008$\pm$0.002 \\
 & CAGCN          &40.9$\pm$3.3  &\cellcolor{gray!20}95.6$\pm$0.2  &0.439$\pm$0.081  &\cellcolor{gray!20}0.037$\pm$0.019 &67.4$\pm$4.1 &\cellcolor{gray!20}95.8$\pm$0.7 &0.255$\pm$0.031 &\cellcolor{gray!20}0.042$\pm$0.022 \\
 & Dirichlet      &42.1$\pm$4.4  &\cellcolor{gray!20}95.5$\pm$0.1  &0.412$\pm$0.039  &\cellcolor{gray!20}0.012$\pm$0.001 &65.9$\pm$3.0  &\cellcolor{gray!20}96.6$\pm$0.4 &0.174$\pm$0.024 &\cellcolor{gray!20}0.009$\pm$0.003\\
 & ETS            &42.1$\pm$5.7  &\cellcolor{gray!20}95.6$\pm$0.2  &0.419$\pm$0.055  &\cellcolor{gray!20}0.009$\pm$0.001 &67.0$\pm$3.9 &\cellcolor{gray!20}95.2$\pm$1.0 &0.176$\pm$0.017 & \cellcolor{gray!20}0.011$\pm$0.003\\
 & GATS           &42.5$\pm$6.2  &\cellcolor{gray!20}95.4$\pm$0.3  &0.433$\pm$0.085  &\cellcolor{gray!20}0.051$\pm$0.020 &66.9$\pm$4.2 &\cellcolor{gray!20}96.5$\pm$0.5 &0.262$\pm$0.042 &\cellcolor{gray!20}0.061$\pm$0.039   \\
 & IRM            &41.8$\pm$4.4  &\cellcolor{gray!20}95.4$\pm$0.2  &0.434$\pm$0.045  &\cellcolor{gray!20}0.009$\pm$0.001 &67.4$\pm$4.1 &\cellcolor{gray!20}96.6$\pm$0.3 & 0.188$\pm$0.021 &\cellcolor{gray!20}0.008$\pm$0.001\\
 & Order &41.1$\pm$3.9  &\cellcolor{gray!20}95.5$\pm$0.1  &0.407$\pm$0.031  &\cellcolor{gray!20}0.012$\pm$0.005 &66.6$\pm$3.8 &\cellcolor{gray!20}96.4$\pm$0.3 &0.179$\pm$0.025 &\cellcolor{gray!20}0.009$\pm$0.002 \\
 & Spline         &41.4$\pm$4.1  &\cellcolor{gray!20}95.5$\pm$0.2  &0.427$\pm$0.043  &\cellcolor{gray!20}0.286$\pm$0.003 &66.1$\pm$2.7 &\cellcolor{gray!20}96.1$\pm$0.8 &0.178$\pm$0.022 &\cellcolor{gray!20}0.287$\pm$0.016 \\
 & VS             &42.7$\pm$5.8  &\cellcolor{gray!20}95.4$\pm$0.3  &0.419$\pm$0.040  &\cellcolor{gray!20}0.007$\pm$0.001 &67.0$\pm$3.9 &\cellcolor{gray!20}96.8$\pm$0.5 &0.176$\pm$0.027  &\cellcolor{gray!20}0.007$\pm$0.003  \\
\midrule

\multirow{8}{*}{\rotatebox{90}{\textbf{Citeseer}}}
& \ding{55}& 46.6$\pm$1.9 &\cellcolor{gray!20}91.2$\pm$0.4&  0.334$\pm$0.027&\cellcolor{gray!20}0.024$\pm$0.004 & 46.6$\pm$1.9&\cellcolor{gray!20}93.4$\pm$0.2  &  0.334$\pm$0.027&\cellcolor{gray!20}0.028$\pm$0.003  \\
 & CAGCN          &46.9$\pm$2.1  &\cellcolor{gray!20}90.9$\pm$0.2  &0.448$\pm$0.057  &\cellcolor{gray!20}0.036$\pm$0.008 &66.5$\pm$3.2 &\cellcolor{gray!20}92.7$\pm$0.2 &0.257$\pm$0.027 &\cellcolor{gray!20}0.041$\pm$0.008 \\
 & Dirichlet      &47.1$\pm$2.0  &\cellcolor{gray!20}90.9$\pm$0.3  &0.333$\pm$0.028  &\cellcolor{gray!20}0.025$\pm$0.002 &66.3$\pm$3.1 &\cellcolor{gray!20}92.8$\pm$0.5 &0.123$\pm$0.023 &\cellcolor{gray!20}0.029$\pm$0.006  \\
 & ETS            &46.7$\pm$1.9  &\cellcolor{gray!20}90.6$\pm$0.4  &0.335$\pm$0.026 &\cellcolor{gray!20}0.025$\pm$0.004 &66.4$\pm$3.3 &\cellcolor{gray!20}93.1$\pm$0.5 &0.132$\pm$0.025 &\cellcolor{gray!20}0.036$\pm$0.004  \\
 & GATS           &47.0$\pm$2.1  &\cellcolor{gray!20}90.9$\pm$0.2  &0.459$\pm$0.050  &\cellcolor{gray!20}0.072$\pm$0.011 &67.1$\pm$3.4 &\cellcolor{gray!20}93.3$\pm$0.3 &0.293$\pm$0.031 &\cellcolor{gray!20}0.052$\pm$0.003    \\
 & IRM            &46.7$\pm$1.9  &\cellcolor{gray!20}90.7$\pm$0.2  &0.330$\pm$0.029  &\cellcolor{gray!20}0.027$\pm$0.003 &66.6$\pm$3.2 &\cellcolor{gray!20}92.7$\pm$0.6 &0.126$\pm$0.027 &\cellcolor{gray!20}0.024$\pm$0.002  \\
 & Order &47.2$\pm$2.5  &\cellcolor{gray!20}90.9$\pm$0.2  &0.317$\pm$0.024  &\cellcolor{gray!20}0.024$\pm$0.004 &67.2$\pm$3.3 &\cellcolor{gray!20}93.1$\pm$0.1 &0.126$\pm$0.014 &\cellcolor{gray!20}0.027$\pm$0.003\\
 & Spline         &46.8$\pm$2.0  &\cellcolor{gray!20}90.9$\pm$0.1  &0.315$\pm$0.026  &\cellcolor{gray!20}0.173$\pm$0.002  &66.3$\pm$3.0 &\cellcolor{gray!20}92.7$\pm$0.7 &0.125$\pm$0.029 &\cellcolor{gray!20}0.177$\pm$0.002 \\
 & VS             &46.9$\pm$2.1  &\cellcolor{gray!20}90.7$\pm$0.4  &0.334$\pm$0.035  &\cellcolor{gray!20}0.021$\pm$0.002  &66.2$\pm$2.9 &\cellcolor{gray!20}92.8$\pm$0.1 &0.125$\pm$0.020 &\cellcolor{gray!20}0.024$\pm$0.003\\
\midrule

\multirow{8}{*}{\rotatebox{90}{\textbf{Pubmed}}}
& \ding{55}& 73.7$\pm$1.9 &\cellcolor{gray!20}90.1$\pm$0.2&  0.146$\pm$0.028&\cellcolor{gray!20}0.006$\pm$0.001 & 73.7$\pm$1.9&\cellcolor{gray!20}85.7$\pm$0.3 &  0.146$\pm$0.028&\cellcolor{gray!20} 0.011$\pm$0.002 \\
 & CAGCN          &74.8$\pm$1.9  &\cellcolor{gray!20}90.1$\pm$0.2  &0.074$\pm$0.017  &\cellcolor{gray!20}0.037$\pm$0.013 &77.4$\pm$2.5 &\cellcolor{gray!20}85.6$\pm$0.2 &0.113$\pm$0.043 &\cellcolor{gray!20}0.052$\pm$0.022  \\
 & Dirichlet      &74.2$\pm$1.6  &\cellcolor{gray!20}90.3$\pm$0.2  &0.134$\pm$0.024  &\cellcolor{gray!20}0.009$\pm$0.001 &77.0$\pm$1.8 &\cellcolor{gray!20}85.6$\pm$0.1 &0.106$\pm$0.016 &\cellcolor{gray!20}0.017$\pm$0.003 \\
 & ETS            &74.9$\pm$1.9  &\cellcolor{gray!20}90.2$\pm$0.2  &0.125$\pm$0.029 &\cellcolor{gray!20}0.007$\pm$0.001 &77.2$\pm$2.3 &\cellcolor{gray!20}85.8$\pm$0.4 &0.104$\pm$0.021 &\cellcolor{gray!20}0.010$\pm$0.003 \\
 & GATS           &73.8$\pm$1.7  &\cellcolor{gray!20}90.2$\pm$0.1  &0.118$\pm$0.033  &\cellcolor{gray!20}0.059$\pm$0.019  &77.3$\pm$2.4 &\cellcolor{gray!20}85.9$\pm$0.2 &0.129$\pm$0.025 &\cellcolor{gray!20}0.079$\pm$0.025  \\
 & IRM            &74.2$\pm$1.5  &\cellcolor{gray!20}90.2$\pm$0.2  &0.141$\pm$0.021  &\cellcolor{gray!20}0.005$\pm$0.001 &77.2$\pm$1.9 &\cellcolor{gray!20}85.5$\pm$0.4 &0.114$\pm$0.022 &\cellcolor{gray!20}0.007$\pm$0.002 \\
 & Order &74.1$\pm$1.8  &\cellcolor{gray!20}90.3$\pm$0.2  &0.136$\pm$0.022  &\cellcolor{gray!20}0.007$\pm$0.001 &77.1$\pm$2.0 &\cellcolor{gray!20}86.1$\pm$0.2 &0.114$\pm$0.025 &\cellcolor{gray!20}\cellcolor{gray!20}0.009$\pm$0.001 \\
 & Spline         &73.8$\pm$1.5  &\cellcolor{gray!20}90.2$\pm$0.1  &0.139$\pm$0.021  &\cellcolor{gray!20}0.084$\pm$0.001 &76.7$\pm$1.6 &\cellcolor{gray!20}85.9$\pm$0.3 &0.116$\pm$0.019 &\cellcolor{gray!20}0.085$\pm$0.009 \\
 & VS             &73.7$\pm$1.2  &\cellcolor{gray!20}90.3$\pm$0.1  &0.148$\pm$0.016  &\cellcolor{gray!20}0.006$\pm$0.000 &77.0$\pm$1.9 &\cellcolor{gray!20}86.4$\pm$0.5 &0.112$\pm$0.019 &\cellcolor{gray!20}0.009$\pm$0.001  \\
\midrule

\multirow{8}{*}{\rotatebox{90}{\textbf{Twitch}}}
& \ding{55}& 57.5$\pm$4.1 &\cellcolor{gray!20}60.6$\pm$1.4& 0.046$\pm$0.028 &\cellcolor{gray!20}0.039$\pm$0.009& 57.5$\pm$4.1&\cellcolor{gray!20}60.1$\pm$3.3 &  0.046$\pm$0.028&\cellcolor{gray!20}0.031$\pm$0.027 \\
 & CAGCN          & 54.4$\pm$4.2 &\cellcolor{gray!20}60.4$\pm$1.3 &0.079$\pm$0.004 &\cellcolor{gray!20}0.091$\pm$0.012 &56.0$\pm$5.2 &\cellcolor{gray!20}61.3$\pm$2.2 &0.072$\pm$0.015 &\cellcolor{gray!20}0.066$\pm$0.009  \\
 & Dirichlet      &55.7$\pm$3.9  &\cellcolor{gray!20}60.4$\pm$1.3  &0.016$\pm$0.006  &\cellcolor{gray!20}0.012$\pm$0.002 &55.9$\pm$5.3 &\cellcolor{gray!20}60.4$\pm$3.0 &0.031$\pm$0.017 &\cellcolor{gray!20}0.019$\pm$0.009  \\
 & ETS            &56.3$\pm$3.3  &\cellcolor{gray!20}59.8$\pm$1.6  &0.037$\pm$0.031  &\cellcolor{gray!20}0.030$\pm$0.010 &55.6$\pm$5.6 &\cellcolor{gray!20}60.2$\pm$2.4&0.063$\pm$0.062  &\cellcolor{gray!20}0.023$\pm$0.011  \\
 & GATS           &55.2$\pm$3.8  &\cellcolor{gray!20}60.4$\pm$1.6  &0.050$\pm$0.020  &\cellcolor{gray!20}0.091$\pm$0.018&55.8$\pm$5.4 &\cellcolor{gray!20}61.2$\pm$2.3 &0.035$\pm$0.018 &\cellcolor{gray!20}0.058$\pm$0.011    \\
 & IRM            &55.6$\pm$3.5  &\cellcolor{gray!20}60.4$\pm$1.7  &0.009$\pm$0.001  &\cellcolor{gray!20}0.011$\pm$0.001&55.5$\pm$5.7 &\cellcolor{gray!20}61.4$\pm$1.1 &0.010$\pm$0.001 &\cellcolor{gray!20}0.012$\pm$0.003 \\
 & Order &54.1$\pm$4.4  &\cellcolor{gray!20}60.6$\pm$1.5  &0.024$\pm$0.025  &\cellcolor{gray!20}0.018$\pm$0.007 &55.3$\pm$6.0 &\cellcolor{gray!20}60.6$\pm$3.5 &0.036$\pm$0.011 &\cellcolor{gray!20}0.019$\pm$0.008 \\
 & Spline         &54.8$\pm$4.1  &\cellcolor{gray!20}60.2$\pm$1.4  &0.140$\pm$0.029  &\cellcolor{gray!20}0.026$\pm$0.015 &55.6$\pm$5.6 &\cellcolor{gray!20}61.8$\pm$1.1 &0.099$\pm$0.106 &\cellcolor{gray!20}0.034$\pm$0.024  \\
 & VS             &55.5$\pm$3.8  &\cellcolor{gray!20}60.8$\pm$1.4   &0.040$\pm$0.026 &\cellcolor{gray!20}0.011$\pm$0.002 &55.8$\pm$5.4 &\cellcolor{gray!20}60.7$\pm$2.1  &0.039$\pm$0.011 &\cellcolor{gray!20}0.015$\pm$0.004  \\
\midrule

\multirow{8}{*}{\rotatebox{90}{\textbf{CBAS}}}
& \ding{55}& 69.6$\pm$4.9 &\cellcolor{gray!20}76.4$\pm$2.8& 0.116$\pm$0.007 &\cellcolor{gray!20}0.086$\pm$0.020 & 70.4$\pm$2.7&\cellcolor{gray!20}59.9$\pm$3.0 & 0.112 $\pm$0.022&\cellcolor{gray!20}0.114$\pm$0.039  \\
 & CAGCN       &65.4$\pm$3.6&72.2$\pm$1.9\cellcolor{gray!20}&0.115$\pm$0.036&0.124$\pm$0.022\cellcolor{gray!20}&66.0$\pm$7.4&57.8$\pm$2.6\cellcolor{gray!20}&0.125$\pm$0.038&0.165$\pm$0.017\cellcolor{gray!20} \\
 & Dirichlet      &67.8$\pm$3.0&76.7$\pm$1.8\cellcolor{gray!20}&0.090$\pm$0.013&0.092$\pm$0.019\cellcolor{gray!20}&66.0$\pm$7.4&61.8$\pm$2.4\cellcolor{gray!20}&0.085$\pm$0.017&0.116$\pm$0.024\cellcolor{gray!20} \\
 & ETS            &68.5$\pm$4.1&76.7$\pm$2.1\cellcolor{gray!20}&0.087$\pm$0.026&0.080$\pm$0.012\cellcolor{gray!20}&65.8$\pm$7.2&59.3$\pm$2.6\cellcolor{gray!20}&0.130$\pm$0.055&0.139$\pm$0.037\cellcolor{gray!20} \\
 & GATS          &68.4$\pm$4.4&78.5$\pm$1.3\cellcolor{gray!20}&0.149$\pm$0.090&0.135$\pm$0.029\cellcolor{gray!20}&65.7$\pm$7.1&62.3$\pm$3.3\cellcolor{gray!20}&0.131$\pm$0.039&0.129$\pm$0.025\cellcolor{gray!20} \\
 & IRM            &68.7$\pm$4.3&76.7$\pm$2.8\cellcolor{gray!20}&0.088$\pm$0.009&0.102$\pm$0.020\cellcolor{gray!20}&63.1$\pm$6.3&60.5$\pm$1.3\cellcolor{gray!20}&0.099$\pm$0.039&0.130$\pm$0.027\cellcolor{gray!20}\\
 & Order &66.3$\pm$5.1&77.1$\pm$1.8\cellcolor{gray!20}&0.069$\pm$0.020&0.098$\pm$0.016\cellcolor{gray!20}&65.2$\pm$6.8&59.3$\pm$2.5\cellcolor{gray!20}&0.118$\pm$0.053&0.129$\pm$0.020\cellcolor{gray!20}\\
 & Spline         &68.7$\pm$4.4&76.7$\pm$2.8\cellcolor{gray!20}&0.136$\pm$0.099&0.126$\pm$0.015\cellcolor{gray!20}&67.1$\pm$8.8&58.7$\pm$2.7\cellcolor{gray!20}&0.152$\pm$0.058&0.143$\pm$0.030\cellcolor{gray!20} \\
 & VS            &67.5$\pm$3.3&74.8$\pm$2.6\cellcolor{gray!20}&0.101$\pm$0.029&0.097$\pm$0.009\cellcolor{gray!20}&69.2$\pm$7.4&58.4$\pm$1.7\cellcolor{gray!20}&0.095$\pm$0.011&0.094$\pm$0.024\cellcolor{gray!20}\\
\bottomrule
\end{tabular}

\end{table*}
\paragraph{Baselines.}
We adopt GCN~\cite{kipf2017semi} and GAT~\cite{velivckovic2018graph} as backbones since they form the foundational frameworks of most GNN models and are widely recognized as important and representative architectures in graph learning. In addition, we consider G-$\Delta$UQ~\cite{trivedi2024accurate}, a recently proposed framework for uncertainty quantification on graphs. G-$\Delta$UQ introduces graph-specific anchoring strategies that regularise the predictive distribution, thereby improving intrinsic uncertainty estimates. Unlike standard post-hoc calibration, it integrates uncertainty modelling directly into the training objective, making it a strong representative of training-based approaches. Finally, we also demonstrate that the combination of our framework with different post-hoc calibration models can further enhance calibration.

In our experiments, we adopt the post-hoc calibration models following \cite{trivedi2024accurate}. Here is an introduction to these post-hoc strategies used in our experiments:
\begin{itemize}[leftmargin=1.2em, itemsep=2pt, topsep=2pt]
    \item CaGCN~\cite{wang2021confident} leverages the graph structure and an auxiliary GCN to generate node-wise temperatures.
    \item Dirichlet calibration~\cite{kull2019beyond} models calibrated outputs with a Dirichlet distribution, capturing inter-class dependency in probability adjustment.
    \item Ensemble temperature scaling (ETS)~\cite{zhang2020mix} extends this idea by combining multiple temperature-scaled models for improved flexibility.  
    \item GATS~\cite{hsu2022makes} further incorporates graph attention to capture the influence of neighboring nodes when learning these temperatures.
    \item Multi-class isotonic regression (IRM)~\cite{zhang2020mix} applies non-parametric isotonic regression to better capture non-linear calibration mappings. 
    \item Order-invariant calibration~\cite{rahimi2020intra} enforces invariance to label permutations, ensuring consistent probability estimates across classes. 
    \item Spline~\cite{gupta2021calibration} fits smooth spline functions to adjust predicted probabilities.  
    \item Vector scaling (VS)~\cite{guo2017calibration} learns class-specific scaling parameters, allowing heterogeneous calibration across classes. 
\end{itemize}

\paragraph{Evaluation Metrics.}
We evaluate performance using accuracy alongside four complementary calibration-oriented metrics: expected calibration error (ECE)~\cite{guo2017calibration}, negative log-likelihood (NLL), Brier Score (BS), and area under the receiver operating characteristic curve (AUROC). Specifically, ECE measures the discrepancy between predicted confidence and empirical accuracy, reflecting the overall calibration quality; NLL assesses how well the predicted probability distribution aligns with the true labels, penalizing overconfident wrong predictions; Brier Score quantifies the mean squared difference between predicted probabilities and actual outcomes, combining both calibration and refinement aspects; and AUROC evaluates the model’s ability to distinguish between positive and negative samples, serving as a threshold-independent indicator of discriminative reliability. Details of how the metrics are calculated are introduced in Appendix~\ref{app:metrics}.

\textbf{Implementation.} SIGHT is implemented with three layers of spiking graph convolution, each followed by predictive coding inference with $K=20$ iterations. Input node features are encoded into Poisson spike trains with $T=25$ time steps, and hidden dimensions are set to 128–128–64. For Cora, Citeseer, and PubMed, we use 500 training epochs with early stopping (patience = 100); for CBAS and Twitch, we use 1000 epochs with patience = 200. The learning rates are set to $\eta_x = 0.005$ and $\eta_p = 0.0005$ for Cora, Citeseer, PubMed, and CBAS, while higher rates $\eta_x = 0.01$ and $\eta_p = 0.001$ are used for Twitch. Each experiment is repeated with five random seeds for statistical robustness. 

All experiments are conducted on a cloud server equipped with a single NVIDIA vGPU (48 GB memory) and 20 vCPUs (Intel Xeon Platinum 8470), with 90 GB system memory. The software environment includes Ubuntu 20.04, Python 3.8, PyTorch 2.0, PyTorch Geometric, snnTorch, and CUDA 11.8. We train models in a full-batch setting with Adam optimizer. For fair comparison, our framework and all baselines employ temperature scaling as a post-processing step for calibration.
As for the ROC thresholds $\tau$, they are not manually chosen but derived from the empirical distribution of uncertainty scores. Formally, let
\begin{equation}
    \mathcal{S} = \{s_1, s_2, \dots, s_n\},
\end{equation}
then the threshold set is $
    \mathcal{T} = \{-\infty\} \cup \mathcal{S} \cup \{+\infty\}$.
That is, each observed uncertainty score is treated as a potential threshold, so AUROC evaluates performance across all possible thresholds rather than relying on a fixed one.

\begin{figure*}[t]
    \centering
    \begin{subfigure}{0.32\textwidth}
        \centering
        \includegraphics[width=\linewidth]{ 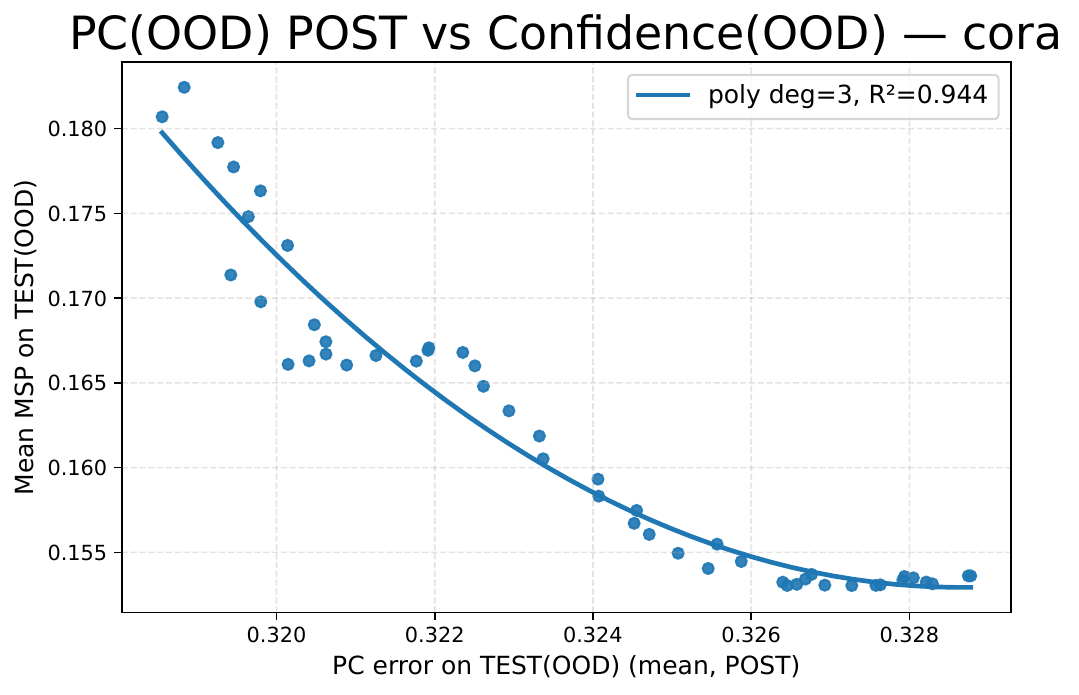} 
        \caption{Cora}
        \label{fig:sub1}
    \end{subfigure}\hfill
    \begin{subfigure}{0.32\textwidth}
        \centering
        \includegraphics[width=\linewidth]{ 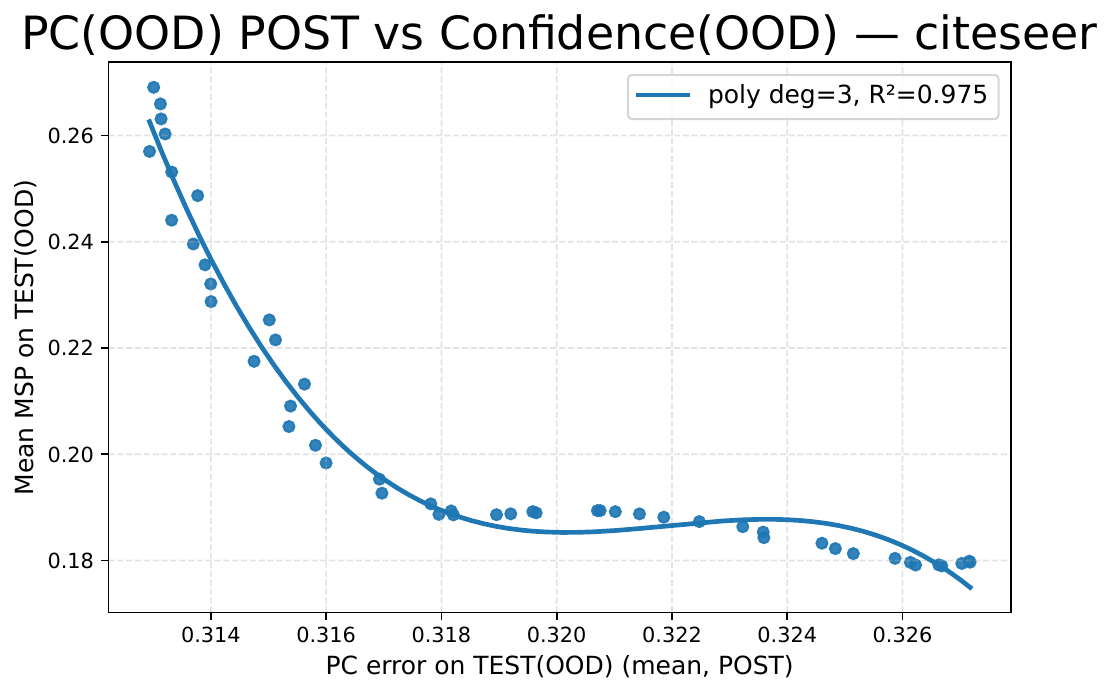}
        \caption{Citeseer}
        \label{fig:sub2}
    \end{subfigure}\hfill
    \begin{subfigure}{0.32\textwidth}
        \centering
        \includegraphics[width=\linewidth]{ 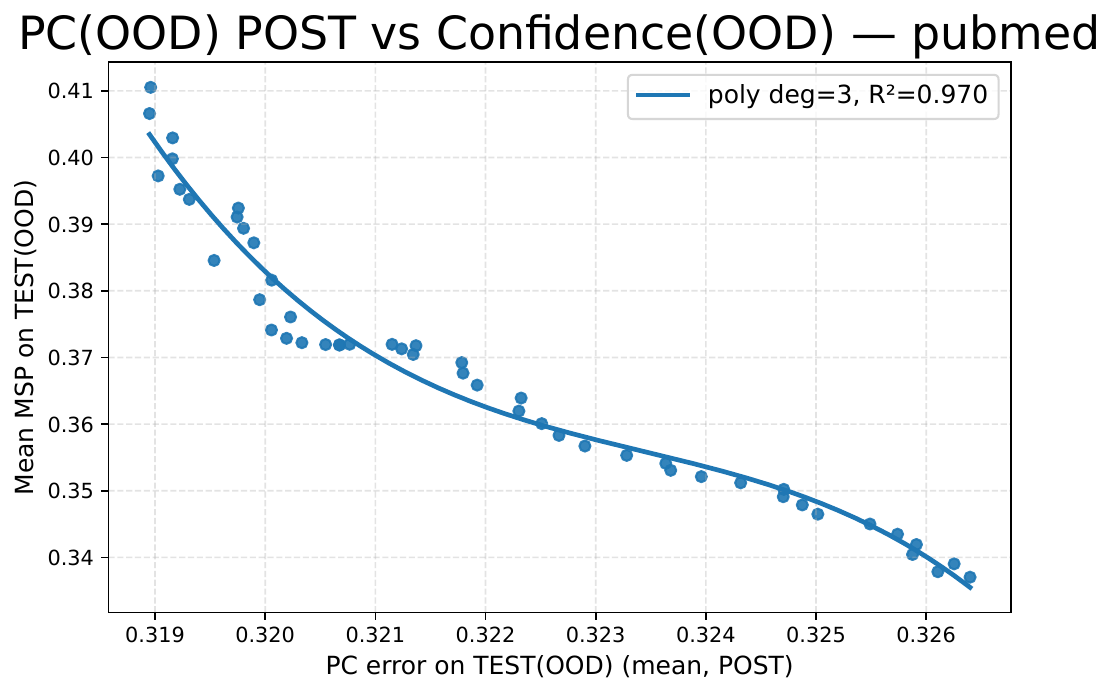}
        \caption{Pubmed}
        \label{fig:sub3}
    \end{subfigure}
    \caption{Correlation between PC error and confidence scores on the Cora, Citeseer, and Pubmed OOD sets. Higher PC errors consistently correspond to lower confidence, showing that PC residuals serve as an intrinsic measure of uncertainty.}
    \label{fig:pc}
\end{figure*}
\subsection{Results}\label{sec:exre}
\paragraph{Comparison of Performance and Calibration.} Table~\ref{tab:PerformanceGCN} compares the node classification accuracy and calibration capabilities of the traditional GNN model on ID and OOD data after introducing different intrinsic calibration methods: G-$\Delta$UQ and SIGHT. Overall, SIGHT achieves the best performance on the majority of benchmarks, particularly in OOD regimes. For example, on Cora and Citeseer, SIGHT significantly improves OOD accuracy by even 40\% while also reducing ECE by 40 times. On Pubmed, SIGHT attains balanced gains, achieving strong ID accuracy and the best results on all OOD scenarios, indicating superior uncertainty awareness. On Twitch and CBAS, SIGHT consistently achieves lower calibration errors together with higher accuracy and AUROC, confirming its strong generalization and uncertainty estimation capabilities under different OOD conditions. These results highlight that \textit{SIGHT not only matches or surpasses accuracy-oriented baselines but also establishes new state-of-the-art calibration and OOD Generalization performance.}

When combining SIGHT with different post-hoc calibration models, as shown in Table~\ref{tab:calibration}, SIGHT consistently produces better-calibrated models, with Accuracy and ECE values outperforming those of the vanilla (w/o SIGHT) models in all datasets. For Cora and Citeseer, incorporating SIGHT into GNNs can increase even more than 50\% Accuracy and ECE can be reduced even to 60 times that without SIGHT. Furthermore, integrating SIGHT with post-hoc calibration models enhances performance more effectively than applying the same calibration strategy to vanilla models. For a given post-hoc method, SIGHT not only improves calibration but also maintains or even surpasses the accuracy of baseline models across most datasets. Although certain combinations of SIGHT and post-hoc calibration methods may lead to slight performance degradation (e.g., SIGHT with GAT on CBAS), this should not be viewed as a limitation of SIGHT itself. Other post-hoc methods applied in the same setting also fail to yield significant improvements in either accuracy or calibration, and in some cases even cause further degradation. In contrast, when combined with GCN, SIGHT already achieves the best performance on CBAS. This suggests that the benefits of SIGHT are more effectively realised through appropriate backbone selection rather than additional post-hoc calibration, highlighting its inherent ability to deliver robust Generalization and well-calibrated uncertainty estimates. Overall, \textit{SIGHT consistently yields better-calibrated models for node classification tasks and can be combined with post-hoc calibration models for further gains.}

\begin{table}
    \centering
    
    \caption{AUROC results for OOD detection. Best results are in \textbf{bold}, and SIGHT is marked with \colorbox{gray!20}{\hspace{1em}\vphantom{a}}.}
    \label{tab:oodd}
    \setlength{\tabcolsep}{3pt}
    \renewcommand{\arraystretch}{0.925}
    \begin{tabular}{lccccc}
        \toprule
        \textbf{Method} & \textbf{Cora} & \textbf{Citeseer} & \textbf{Pubmed} & \textbf{Twitch}  & \textbf{CBAS}  \\
        \midrule
        GCN &70.4$\pm$2.8  &74.1$\pm$1.1  &72.6$\pm$1.2  &53.2$\pm$0.5  & 74.5$\pm$3.8 \\
        ~+G-$\Delta$UQ &83.0$\pm$2.2  &77.7$\pm$1.0  &74.8$\pm$0.4  &52.2$\pm$4.6  &  72.3$\pm$6.6 \\
        \rowcolor{gray!20}~+SIGHT & \textbf{89.0$\pm$1.6}  &\textbf{84.0$\pm$1.2}  &\textbf{85.1$\pm$0.2}  &\textbf{54.5$\pm$2.4}    &  78.1$\pm$2.4\\
        \midrule
        GAT &73.9$\pm$2.1  &78.3$\pm$1.1  &77.0$\pm$0.6  &49.5$\pm$2.4    & 75.9$\pm$4.1\\
        ~+G-$\Delta$UQ &83.3$\pm$2.4  &80.4$\pm$1.1  &82.7$\pm$0.8  &49.3$\pm$3.9  &  78.3$\pm$5.3 \\
        \rowcolor{gray!20}~+SIGHT&88.0$\pm$2.8  &81.1$\pm$1.8  &82.1$\pm$0.2  &52.7$\pm$3.1  & \textbf{83.1$\pm$3.4}\\
        \bottomrule
    \end{tabular}
\end{table}

\begin{table*}[t]
\centering

\caption{Ablation study of SIGHT with GCN and GAT backbones. Best results are in \textbf{bold}.}
\label{tab:AblationBoth}
\setlength{\tabcolsep}{3pt}
\resizebox{\linewidth}{!}{
\renewcommand{\arraystretch}{0.925}
\begin{tabular}{c l
                cc cc cc
                cc cc cc}
\toprule
& & \multicolumn{5}{c}{\textbf{GCN}} & \multicolumn{5}{c}{\textbf{GAT}} \\
\cmidrule(lr){3-7}\cmidrule(lr){8-12}
 & \textbf{Method} 
& \textbf{Acc} $\uparrow$ & \textbf{ECE} $\downarrow$ & \textbf{NLL} $\downarrow$ & \textbf{BS} $\downarrow$ & \textbf{AUROC} $\uparrow$ 
& \textbf{Acc} $\uparrow$ & \textbf{ECE} $\downarrow$ & \textbf{NLL} $\downarrow$ & \textbf{BS} $\downarrow$ & \textbf{AUROC} $\uparrow$ \\
\midrule

\multirow{4}{*}{\rotatebox{90}{\textbf{Cora}}}
& SIGHT         & \textbf{95.6$\pm$0.2} & \textbf{0.009$\pm$0.002} & \textbf{0.153$\pm$0.010} & \textbf{0.068$\pm$0.003} & \textbf{89.8$\pm$1.5} 
                 & 96.3$\pm$0.7 & \textbf{0.008$\pm$0.002} & \textbf{0.138$\pm$0.028} & 0.058$\pm$0.012 & \textbf{89.6$\pm$2.1} \\
& ~w/o Spik   & 68.4$\pm$5.1 & 0.146$\pm$0.043 & 1.011$\pm$0.196 & 0.473$\pm$0.084 & 71.4$\pm$2.3 
                 & 64.5$\pm$5.5 & 0.175$\pm$0.048 & 1.243$\pm$0.230 & 0.551$\pm$0.093 & 66.3$\pm$4.7 \\
& ~w/o PC        & 95.2$\pm$0.3 & 0.020$\pm$0.005 & 0.158$\pm$0.004 & 0.076$\pm$0.002 & 89.3$\pm$1.4 
                 & \textbf{97.1$\pm$0.2} & 0.017$\pm$0.005 & 0.135$\pm$0.022 & \textbf{0.048$\pm$0.006} & 83.3$\pm$5.2 \\
& ~w/o SPC       & 43.5$\pm$5.2 & 0.391$\pm$0.071 & 2.595$\pm$0.512 & 0.919$\pm$0.103 & 59.9$\pm$3.7 
                 & 67.4$\pm$4.1 & 0.164$\pm$0.026 & 1.072$\pm$0.108 & 0.498$\pm$0.059 & 71.1$\pm$3.2 \\
\midrule

\multirow{4}{*}{\rotatebox{90}{\textbf{Citeseer}}}
& SIGHT         & 91.2$\pm$0.4 & \textbf{0.024$\pm$0.004} & 0.307$\pm$0.016 & 0.136$\pm$0.008 & \textbf{83.4$\pm$0.8} 
                 & \textbf{93.4$\pm$0.2} & \textbf{0.028$\pm$0.003} & \textbf{0.261$\pm$0.012} & \textbf{0.105$\pm$0.005} & \textbf{80.8$\pm$1.2} \\
& ~w/o Spik   & 75.0$\pm$2.7 & 0.040$\pm$0.007 & 0.793$\pm$0.081 & 0.372$\pm$0.036 & 75.6$\pm$1.4 
                 & 79.1$\pm$3.6 & 0.038$\pm$0.009 & 0.678$\pm$0.096 & 0.315$\pm$0.047 & 78.3$\pm$1.3 \\
& ~w/o PC        & \textbf{92.1$\pm$0.2} & 0.037$\pm$0.005 & \textbf{0.296$\pm$0.011} & \textbf{0.126$\pm$0.004} & 82.3$\pm$1.3 
                 & 93.2$\pm$0.3 & 0.051$\pm$0.007 & 0.287$\pm$0.009 & 0.110$\pm$0.005 & 79.0$\pm$1.9 \\
& ~w/o SPC       & 46.6$\pm$1.9 & 0.334$\pm$0.027 & 1.838$\pm$0.051 & 0.793$\pm$0.026 & 74.1$\pm$1.6 
                 & 66.6$\pm$3.3 & 0.125$\pm$0.022 & 0.968$\pm$0.085 & 0.470$\pm$0.037 & 77.4$\pm$1.3 \\
\midrule

\multirow{4}{*}{\rotatebox{90}{\textbf{Pubmed}}}
& SIGHT         & \textbf{90.1$\pm$0.2} & \textbf{0.006$\pm$0.001} & \textbf{0.275$\pm$0.004} & \textbf{0.148$\pm$0.002} & \textbf{85.1$\pm$0.2} 
                 & 85.7$\pm$0.3 & 0.011$\pm$0.002 & 0.380$\pm$0.009 & 0.210$\pm$0.005 & 81.7$\pm$0.8 \\
& ~w/o Spik   & 78.3$\pm$1.2 & 0.116$\pm$0.013 & 0.665$\pm$0.043 & 0.338$\pm$0.018 & 77.1$\pm$1.6 
                 & 79.1$\pm$3.6 & 0.038$\pm$0.009 & 0.678$\pm$0.096 & 0.315$\pm$0.047 & 78.3$\pm$1.3 \\
& ~w/o PC        & 87.2$\pm$0.1 & 0.008$\pm$0.001 & 0.345$\pm$0.002 & 0.189$\pm$0.001 & 81.8$\pm$0.1 
                 & \textbf{92.9$\pm$0.2} & \textbf{0.008$\pm$0.001} & \textbf{0.203$\pm$0.006} & \textbf{0.107$\pm$0.003} & \textbf{87.3$\pm$0.4} \\
& ~w/o SPC       & 73.7$\pm$1.9 & 0.146$\pm$0.028 & 0.960$\pm$0.081 & 0.411$\pm$0.028 & 72.3$\pm$1.5 
                 & 76.7$\pm$1.5 & 0.116$\pm$0.018 & 0.776$\pm$0.078 & 0.352$\pm$0.022 & 76.0$\pm$0.7 \\
\midrule

\multirow{4}{*}{\rotatebox{90}{\textbf{Twitch}}}
& SIGHT         & \textbf{60.6$\pm$1.4} & 0.039$\pm$0.009 & \textbf{0.671$\pm$0.005} & 0.478$\pm$0.005 & \textbf{52.2$\pm$3.9} 
                 & \textbf{60.1$\pm$3.3} & \textbf{0.031$\pm$0.027} & \textbf{0.670$\pm$0.013} & \textbf{0.477$\pm$0.012} & \textbf{51.6$\pm$4.7} \\
& ~w/o Spik   & 59.5$\pm$5.5 & 0.016$\pm$0.003 & 0.669$\pm$0.012 & \textbf{0.476$\pm$0.012} & 50.2$\pm$4.5 
                 & 55.4$\pm$7.9 & 0.072$\pm$0.041 & 0.681$\pm$0.014 & 0.488$\pm$0.014 & 46.6$\pm$7.6 \\
& ~w/o PC        & 59.6$\pm$2.2 & \textbf{0.035$\pm$0.021} & 0.675$\pm$0.011 & 0.482$\pm$0.010 & 51.8$\pm$4.1 
                 & 59.6$\pm$2.9 & 0.038$\pm$0.029 & 0.673$\pm$0.012 & 0.481$\pm$0.012 & 51.3$\pm$5.1 \\
& ~w/o SPC       & 57.5$\pm$4.1 & 0.046$\pm$0.028 & 0.683$\pm$0.007 & 0.490$\pm$0.008 & 49.2$\pm$3.3 
                 & 58.6$\pm$3.7 & 0.036$\pm$0.017 & 0.678$\pm$0.012 & 0.485$\pm$0.012 & 49.7$\pm$2.6 \\
\midrule

\multirow{4}{*}{\rotatebox{90}{\textbf{CBAS}}}
& SIGHT        &\textbf{76.4$\pm$2.8}    &\textbf{0.086$\pm$0.020}  &0.746$\pm$0.075  &0.363$\pm$0.032 &73.0$\pm$1.8 &60.0$\pm$3.0&0.114$\pm$0.039\textbf{}& 0.978$\pm$0.054\textbf{}&0.504$\pm$0.029\textbf{} &\textbf{81.0$\pm$5.3}\\
& ~w/o Spik   &74.8$\pm$6.2 &0.093$\pm$0.021 &0.726$\pm$0.105 &0.359$\pm$0.066 &73.8$\pm$3.5 
                 & \textbf{73.0$\pm$2.8 }& \textbf{0.088$\pm$0.019} & \textbf{0.802$\pm$0.030} & \textbf{0.396$\pm$0.008 }& 72.3$\pm$6.7 \\
& ~w/o PC        &76.0$\pm$3.8 & 0.097$\pm$0.009 & \textbf{0.645$\pm$0.115} & \textbf{0.334$\pm$0.061} &\textbf{82.8$\pm$7.3}
                 & 70.4$\pm$3.1 & 0.122$\pm$0.033 & 0.839$\pm$0.046 & 0.427$\pm$0.025 & 76.0$\pm$4.9 \\
& ~w/o SPC       & 69.6$\pm$4.9   & 0.116$\pm$0.007 & 0.840$\pm$0.026 &0.432$\pm$0.022& 71.6$\pm$2.6 
                 &70.4$\pm$2.6& 0.112$\pm$0.022 & 0.836$\pm$0.074 & 0.429$\pm$0.039 & 70.4$\pm$2.8\\
\bottomrule
\end{tabular}
}
\end{table*}

\paragraph{OOD Detection.}
OOD detection~\cite{hendrycks2019deep,ren2023graph} aims to classify samples as ID or OOD. To evaluate whether models assign higher uncertainty to shifted inputs~\cite{bazhenov2023evaluating}, we compare G-$\Delta$UQ and SIGHT with GCN and GAT backbones. Their AUROC results are reported in Table~\ref{tab:oodd}.

Across the three citation networks and the social network Twitch, SIGHT consistently achieves the highest AUROC under GCN backbone, substantially outperforming plain GCN and even the strong G-$\Delta$UQ baseline nearly 20\% and 7\%  at most, respectively. For the CBAS dataset, SIGHT with GAT remains competitive and surpasses the baselines, confirming its trustworthiness under diverse distribution shifts. These improvements highlight the effectiveness of predictive coding residuals in distinguishing ID from OOD samples.  Overall, these results demonstrate that \textit{SIGHT provides reliable uncertainty estimation for OOD detection, outperforming both conventional GNNs and advanced calibration baselines.}

\paragraph{Explainability of Predictive Coding on Uncertainty Estimation.}
This experiment measures PC error as an interpretability signal in our model trained on OOD datasets. Figure~\ref{fig:pc} illustrates the correlation between PC error and model confidence measured by the mean maximum softmax probability (MSP), on the OOD splits of Cora, Citeseer, and Pubmed. 
Across all three datasets, we observe a strong negative correlation: as the average PC error increases, the MSP decreases. 
This indicates that larger residuals correspond to lower confidence, thereby providing an intrinsic and interpretable signal of uncertainty. 
Notably, the fitted polynomial curves show high $R^2$ values ($>0.94$), confirming that PC errors serve as a reliable proxy for confidence across diverse graph benchmarks. 
These results highlight that \textit{SGPC does not require additional calibration procedures to extract meaningful uncertainty estimates, as the residuals themselves naturally align with confidence scores under distribution shifts.}

\paragraph{Ablation Study}
Table~\ref{tab:AblationBoth} reports an ablation study with GCN and GAT backbone, where we 
compare the full SIGHT model against three variants: without spiking neurons, 
without predictive coding, and without the joint Spiking Predictive Coding mechanism. 
The complete SIGHT model consistently achieves the best performance across all datasets and metrics, confirming the effectiveness of combining spiking computation with predictive coding. 
When the spiking mechanism is removed, accuracy drops substantially and calibration metrics such as ECE and Brier Score deteriorate, indicating that event-driven representations are key to both Generalization and calibration. 
Excluding predictive coding yields relatively high accuracy on some datasets, but calibration performance significantly worsens, as reflected by higher NLL and ECE values, showing that local error feedback is essential for uncertainty estimation and well-calibrated predictions. 
Finally, removing the full spiking predictive coding loop leads to the most severe degradation across metrics. These results collectively demonstrate that \textit{both spiking computation and predictive coding contribute complementary strengths.
Their integration in SIGHT is thus crucial for delivering high accuracy, strong calibration, and reliable OOD generalization.}

From the perspective of dataset characteristics, it can also be observed that incorporating SIGHT into the same backbone consistently yields strong performance across different types of distribution shifts. Specifically, integrating SIGHT with GCN leads to better results on Cora, Pubmed, and CBAS, while combining it with GAT achieves the best performance on Citeseer and Twitch. 


\vspace{-0.5em} 
\section{Conclusion}
In this work, we developed a plug-in module designed to strengthen uncertainty-aware and interpretable graph learning in high-stakes social systems, where reliability and transparency is essential for responsible decision-making. SIGHT departs from deterministic GNNs by iteratively minimizing prediction errors over spiking graph states, enabling internal mismatch signals to surface when the inference process becomes unstable. This error-driven mechanism offers intuitive and principled interpretability for uncertainty. By revealing uncertainty through transparent internal dynamics rather than opaque confidence scores, SIGHT supports safer and more accountable decision-making in environments where model failures can have significant societal consequences. Extensive experiments demonstrate that SIGHT consistently improves uncertainty estimation and transparency across diverse distribution shifts while maintaining competitive predictive performance. These results highlight SIGHT as a promising direction for building trustworthy graph learning models tailored to socially critical applications.

For future work, we plan to extend SIGHT by disentangling its prediction–observation mismatch signals into node-, edge-, and structure-level components, enabling users to identify whether uncertainty originates from noisy features, unreliable neighbors, or distribution drift. 


\bibliographystyle{ACM-Reference-Format}
\balance
\bibliography{sample-base}

\appendix

\section{Experiments}
\label{app:experiments}

\subsection{Datasets}
In this section, we provide a detailed analysis of the distributional shifts present in our benchmark datasets. We examine how source and target graphs differ in terms of feature distributions and label semantics, and quantify the extent to which these variations manifest as covariate shift and concept shift.





\subsection{Evaluation Metrics}
\label{app:metrics}

According to section~\ref{subsec:exsetup}, we report five widely used metrics: Accuracy, Expected Calibration Error (ECE), Negative Log-Likelihood (NLL), Brier Score (BS), and the Area Under the Receiver Operating Characteristic Curve (AUROC). These metrics capture complementary aspects of accuracy and calibration.

\paragraph{Accuracy.}  
Accuracy is the most widely used metric for evaluating node classification performance. It measures the proportion of correctly predicted labels over the total number of test nodes and directly reflects the discriminative power of the model. Formally, given test set $\mathcal{D}_{\text{test}}$ with ground-truth labels $\{y_i\}$ and predicted labels $\{\hat{y}_i\}$, accuracy is defined as
\begin{equation}
    \text{Accuracy} = \frac{1}{|\mathcal{D}_{\text{test}}|} \sum_{i \in \mathcal{D}_{\text{test}}} \mathbb{I}(\hat{y}_i = y_i),
\end{equation}where $\mathbb{I}(\cdot)$ denotes the indicator function. 

\begin{table*}[t]
\centering

\caption{Node classification accuracy and uncertainty calibration on ID and OOD datasets with the GAT backbone. Best results are in \textbf{bold}, and SIGHT is marked with \colorbox{gray!20}{\hspace{1em}\vphantom{a}}. Results with the GCN backbone are in Table~\ref{tab:PerformanceGCN}.}
\label{tab:PerformanceGAT}
\setlength{\tabcolsep}{2.5pt}
\resizebox{\linewidth}{!}{
\begin{tabular}{c l
                cc cc  
                cc cc cc} 
\toprule
\multicolumn{1}{c}{\multirow{2}{*}{}} & \multicolumn{1}{c}{\multirow{2}{*}{\textbf{Method}}}
& \multicolumn{2}{c}{\textbf{Accuracy} $\uparrow$}
& \multicolumn{2}{c}{\textbf{ECE} $\downarrow$}
& \multicolumn{2}{c}{\textbf{NLL} $\downarrow$}
& \multicolumn{2}{c}{\textbf{BS} $\downarrow$}
& \multicolumn{2}{c}{\textbf{AUROC} $\uparrow$}
\\\cmidrule(lr){3-4}\cmidrule(lr){5-6}\cmidrule(lr){7-8}\cmidrule(lr){9-10}\cmidrule(lr){11-12}
& & ID & OOD & ID & OOD & ID & OOD & ID & OOD & ID & OOD \\
\midrule

\multirow{3}{*}{\rotatebox{90}{\textbf{Cora}}}
 & GAT        &90.6$\pm$0.5  &67.4$\pm$4.1  &\textbf{0.015$\pm$0.001}  &0.164$\pm$0.026  &0.286$\pm$0.027  &1.072$\pm$0.108  &0.137$\pm$0.011  &0.498$\pm$0.059  &\textbf{88.9$\pm$0.7} &71.1$\pm$3.2\\
 & ~+G-$\Delta$UQ       &93.9$\pm$0.4 &80.2$\pm$3.5 &0.015$\pm$0.002 &0.061$\pm$0.033 &\textbf{0.211$\pm$0.007} &0.578$\pm$0.098 &\textbf{0.097$\pm$0.004} &0.291$\pm$0.053 &87.2$\pm$1.6 &81.8$\pm$3.0\\  
 & \cellcolor{gray!20}+SIGHT          &\cellcolor{gray!20}\textbf{93.9$\pm$0.9}  &\cellcolor{gray!20}\textbf{96.3$\pm$0.7}  &\cellcolor{gray!20}0.017$\pm$0.002  &\cellcolor{gray!20}\textbf{0.008$\pm$0.002}  &\cellcolor{gray!20}0.227$\pm$0.036  &\cellcolor{gray!20}\textbf{0.138$\pm$0.028}  &\cellcolor{gray!20}0.097$\pm$0.016  &\cellcolor{gray!20}\textbf{0.058$\pm$0.012} &\cellcolor{gray!20}87.1$\pm$3.0 &\cellcolor{gray!20}\textbf{89.6$\pm$2.1}\\
\midrule

\multirow{3}{*}{\rotatebox{90}{\textbf{Citeseer}}}
 & GAT        &82.1$\pm$0.4  &46.6$\pm$1.9  &0.026$\pm$0.003  &0.334$\pm$0.027  &0.503$\pm$0.012 &1.838$\pm$0.051  &0.252$\pm$0.006 &0.793$\pm$0.026&\textbf{84.1$\pm$0.8}&74.1$\pm$1.6  \\
 & ~+G-$\Delta$UQ           &81.6$\pm$0.4  &71.0$\pm$4.2  &0.021$\pm$0.008  &0.066$\pm$0.027  &0.531$\pm$0.037  &0.889$\pm$0.145 &0.263$\pm$0.010 &0.416$\pm$0.056 &82.2$\pm$1.6 &76.5$\pm$1.3\\  
 & \cellcolor{gray!20}+SIGHT           &\cellcolor{gray!20}\textbf{88.8$\pm$0.5}  &\cellcolor{gray!20}\textbf{93.4$\pm$0.2}  &\cellcolor{gray!20}\textbf{0.020$\pm$0.006}  &\cellcolor{gray!20}\textbf{0.028$\pm$0.003}  &\cellcolor{gray!20}\textbf{0.414$\pm$0.022}  &\cellcolor{gray!20}\textbf{0.261$\pm$0.012}  &\cellcolor{gray!20}\textbf{0.175$\pm$0.010}  &\cellcolor{gray!20}\textbf{0.105$\pm$0.005}  &\cellcolor{gray!20}77.7$\pm$1.74 &\cellcolor{gray!20}\textbf{80.8$\pm$1.2}\\ 
\midrule

\multirow{3}{*}{\rotatebox{90}{\textbf{Pubmed}}}

 & GAT        & 88.0$\pm$0.1  &73.7$\pm$1.9  &0.008$\pm$0.001  &0.146$\pm$0.028  &0.306$\pm$0.001&0.960$\pm$0.081  &0.173$\pm$0.000 &0.411$\pm$0.028&85.2$\pm$0.3&72.3$\pm$1.5  \\
 & ~+G-$\Delta$UQ           &\textbf{91.3$\pm$0.2}  &83.8$\pm$0.7  &\textbf{0.006$\pm$0.001}  &0.090$\pm$0.009 &\textbf{0.236$\pm$0.002} &0.668$\pm$0.048 &\textbf{0.129$\pm$0.002} &0.263$\pm$0.013 &\textbf{86.3$\pm$0.4} &74.2$\pm$0.3\\  
 & \cellcolor{gray!20}+SIGHT          &\cellcolor{gray!20}86.0$\pm$0.3  &\cellcolor{gray!20}\textbf{85.7}$\pm$0.3 &\cellcolor{gray!20}0.012$\pm$0.002  &\cellcolor{gray!20}\textbf{0.011$\pm$0.002}  &\cellcolor{gray!20}0.376$\pm$0.009  &\cellcolor{gray!20}\textbf{0.380$\pm$0.009}  &\cellcolor{gray!20}0.208$\pm$0.005  &\cellcolor{gray!20}\textbf{0.210$\pm$0.005}&\cellcolor{gray!20}81.6$\pm$0.7 &\cellcolor{gray!20}\textbf{81.7$\pm$0.8}\\
\midrule

\multirow{3}{*}{\rotatebox{90}{\textbf{Twitch}}}

 & GAT        &60.7$\pm$9.0  &57.5$\pm$4.1  & 0.129$\pm$0.066  &0.046$\pm$0.028  &\textbf{0.685}$\pm$0.006&0.683$\pm$0.007  & \textbf{0.492}$\pm$0.006 &0.490$\pm$0.008&\textbf{59.5$\pm$4.3}& 49.2$\pm$3.3 \\
 & ~+G-$\Delta$UQ           &\textbf{64.8$\pm$9.9}  &58.3$\pm$2.9  &0.163$\pm$0.067  &0.065$\pm$0.024 &0.690$\pm$0.021  &0.687$\pm$0.009& 0.497$\pm$0.021 &0.494$\pm$0.009 &58.0$\pm$9.1 &45.4$\pm$1.5 \\  
 & \cellcolor{gray!20}+SIGHT            &\cellcolor{gray!20}51.2$\pm$2.9  &\cellcolor{gray!20}\textbf{60.1$\pm$3.3}  &\cellcolor{gray!20}\textbf{0.103}$\pm$0.033  & \cellcolor{gray!20}\textbf{0.031$\pm$0.027}  &\cellcolor{gray!20}0.719$\pm$0.021  &\cellcolor{gray!20}\textbf{0.670$\pm$0.013}  &\cellcolor{gray!20}0.524$\pm$0.019  &\cellcolor{gray!20}\textbf{0.477$\pm$0.012}  &\cellcolor{gray!20}49.8$\pm$2.4 &\cellcolor{gray!20}\textbf{51.6$\pm$4.7}\\
\midrule

\multirow{3}{*}{\rotatebox{90}{\textbf{CBAS}}}

 & GAT        &75.7$\pm$1.9&70.4$\pm$2.7&0.104$\pm$0.034&\textbf{0.112$\pm$0.022}&0.593$\pm$0.042&0.837$\pm$0.074&0.324$\pm$0.026&0.429$\pm$0.040&81.5$\pm$6.3&70.4$\pm$2.8 \\
 & ~+G-$\Delta$UQ            &80.1$\pm$6.8&\textbf{73.0$\pm$7.2}&\textbf{0.101$\pm$0.030}&0.123$\pm$0.047&0.529$\pm$0.128&\textbf{0.759$\pm$0.134}&0.284$\pm$0.072&\textbf{0.389$\pm$0.082}&82.1$\pm$2.3&74.8$\pm$8.1 \\     
 & \cellcolor{gray!20}+SIGHT          &\cellcolor{gray!20}\textbf{82.6 $\pm$0.6}&\cellcolor{gray!20}60.0$\pm$3.0&\cellcolor{gray!20}0.126$\pm$0.027\textbf{}&\cellcolor{gray!20}0.114$\pm$0.039\textbf{}&\cellcolor{gray!20}\textbf{0.515$\pm$0.034}&\cellcolor{gray!20}0.978$\pm$0.054\textbf{}&\cellcolor{gray!20}\textbf{0.261$\pm$0.013}&\cellcolor{gray!20}0.504$\pm$0.029\textbf{} &\cellcolor{gray!20}\textbf{84.6$\pm$0.0}&\cellcolor{gray!20}\textbf{81.0$\pm$5.3}\\
\bottomrule
\end{tabular}
}
\end{table*}
\paragraph{Expected Calibration Error (ECE)}  
Calibration refers to the alignment between predicted confidence and empirical accuracy. Calibrated models are expected to generate confidence scores that accurately reflect the true likelihood of the predicted classes~\cite{naeini2015obtaining,guo2017calibration,ovadia2019can}. ECE partitions predictions into $M$ confidence bins $\{B_m\}_{m=1}^M$. For each bin, we compute the average confidence $\text{conf}(B_m)$ and the empirical accuracy $\text{acc}(B_m)$. ECE is defined as
\begin{equation}
    \text{ECE} = \sum_{m=1}^M \frac{|B_m|}{n} \big|\text{acc}(B_m) - \text{conf}(B_m)\big|,
\end{equation} where $n$ is the total number of test samples. A smaller ECE indicates better calibration.

\paragraph{Brier Score (BS).}  
The Brier Score measures the mean squared difference between predicted probability vectors and one-hot ground-truth labels:
\begin{equation}
    \text{Brier} = \frac{1}{n} \sum_{i=1}^n \sum_{c=1}^C \big(p_\theta(y=c|x_i) - \mathbf{1}[y_i=c]\big)^2,
\end{equation} where $C$ is the number of classes and $\mathbf{1}[\cdot]$ denotes the indicator function. A lower Brier Score indicates that predicted probabilities are closer to the true distribution.

\paragraph{Negative Log-Likelihood (NLL)}  
NLL evaluates the quality of probabilistic predictions. Given predicted probability distributions $p_\theta(y_i|x_i)$ for test samples $(x_i,y_i)$, it is defined as
\begin{equation}
    \text{NLL} = -\frac{1}{n} \sum_{i=1}^n \log p_\theta(y_i|x_i).
\end{equation}
Lower values correspond to higher likelihood assigned to the true labels, indicating better uncertainty modelling.

\paragraph{Area Under ROC Curve (AUROC)}  
Given a classifier that outputs class probabilities $\{p_{i,c}\}_{c=1}^K$ (or logits $\{z_{i,c}\}_{c=1}^K$) for each sample $i$, let the hard prediction be
\begin{equation}
    \hat{y}_i=argmax_{c} \, p_{i,c}, \qquad 
y_i \in \{1,\dots,K\}.
\end{equation}
Define the binary label for \emph{error detection} as
\begin{equation}
y^{\text{bin}}_i = \indicator{\hat{y}_i \neq y_i} \in \{0,1\},
\end{equation}
where $1$ denotes a misclassification (positive class) and $0$ denotes a correct prediction (negative class).

We compute an \emph{uncertainty score} $s_i$ from the model outputs using MSP and entropy:
\begin{align}
 \text{MSP:}\quad & s_i = 1 - \max_{c} p_{i,c},\\
\text{Entropy:}\quad & s_i = -\sum_{c=1}^K p_{i,c}\,\log p_{i,c}.
\end{align}
For a threshold $\tau$, predict ``error'' if $s_i>\tau$. The ROC statistics are
\begin{align}
\text{TPR}(\tau) &= 
\frac{\sum_i \indicator{s_i>\tau}\,\indicator{y^{\text{bin}}_i=1}}
     {\sum_i \indicator{y^{\text{bin}}_i=1}},\\
\text{FPR}(\tau) &=
\frac{\sum_i \indicator{s_i>\tau}\,\indicator{y^{\text{bin}}_i=0}}
     {\sum_i \indicator{y^{\text{bin}}_i=0}}.
\end{align}
The \emph{AUROC} is the area under this ROC curve:
\begin{equation}
\text{AUROC} \;=\; \int_{0}^{1} 
\text{TPR}\!\big(\text{FPR}^{-1}(u)\big)\, \mathrm{d}u.
\end{equation}
Intuitively, it measures how well $s_i$ ranks misclassified samples above correctly classified ones. 

\begin{figure*}[t]
  \centering
  \begin{subfigure}{0.32\textwidth}
    \centering
    \includegraphics[width=\linewidth]{ 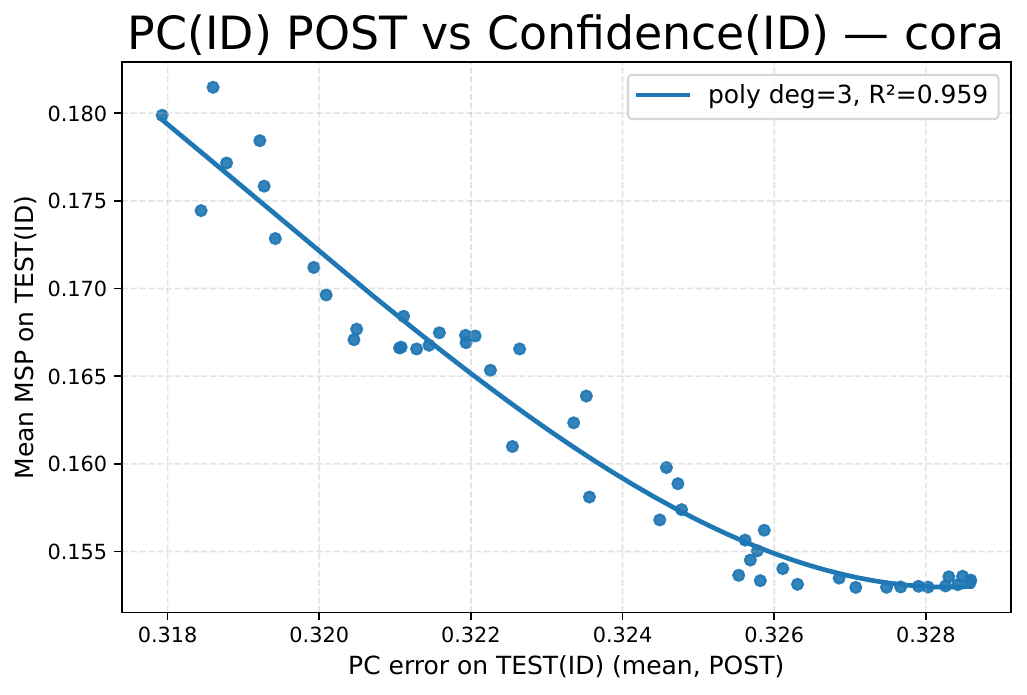} 
    \caption{Cora}
    \label{fig:sub1}
  \end{subfigure}\hfill
  \begin{subfigure}{0.32\textwidth}
    \centering
    \includegraphics[width=\linewidth]{ 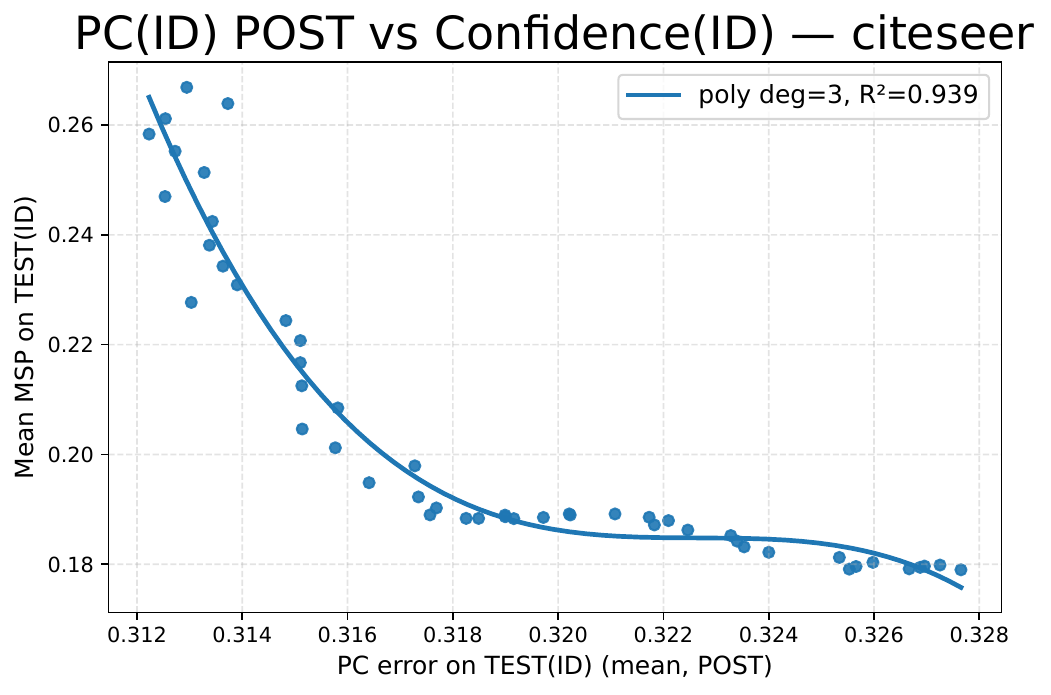}
    \caption{Citeseer}
    \label{fig:sub2}
  \end{subfigure}\hfill
  \begin{subfigure}{0.32\textwidth}
    \centering
    \includegraphics[width=\linewidth]{ 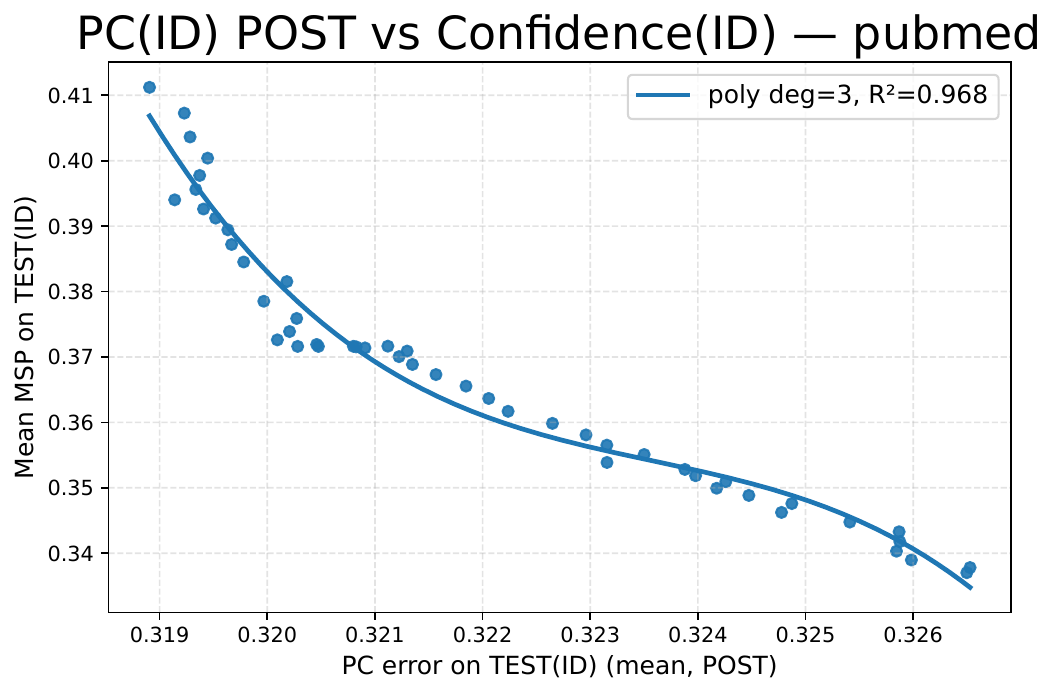}
    \caption{Pubmed}
    \label{fig:sub3}
  \end{subfigure}

    \caption{Correlation between PC error and confidence scores on the Cora, Citeseer, and Pubmed ID test sets. Higher PC errors consistently correspond to lower confidence, showing that PC residuals serve as an intrinsic measure of uncertainty.}
  \label{fig:pcID3}
\end{figure*}


\begin{figure*}[t]
    \centering
    \begin{subfigure}{0.31\textwidth}
        \centering
        \includegraphics[width=\linewidth]{ 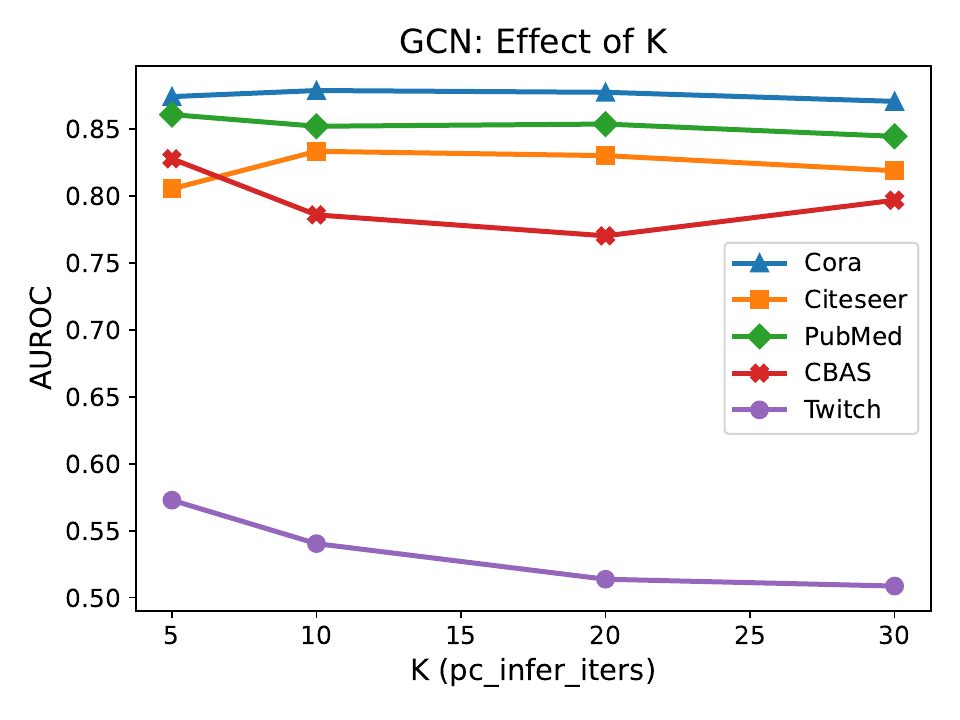}
        \caption{$\text{GCN}_K$}
        \label{fig:sub1}
    \end{subfigure}\hfill
    \begin{subfigure}{0.31\textwidth}
        \centering
        \includegraphics[width=\linewidth]{ 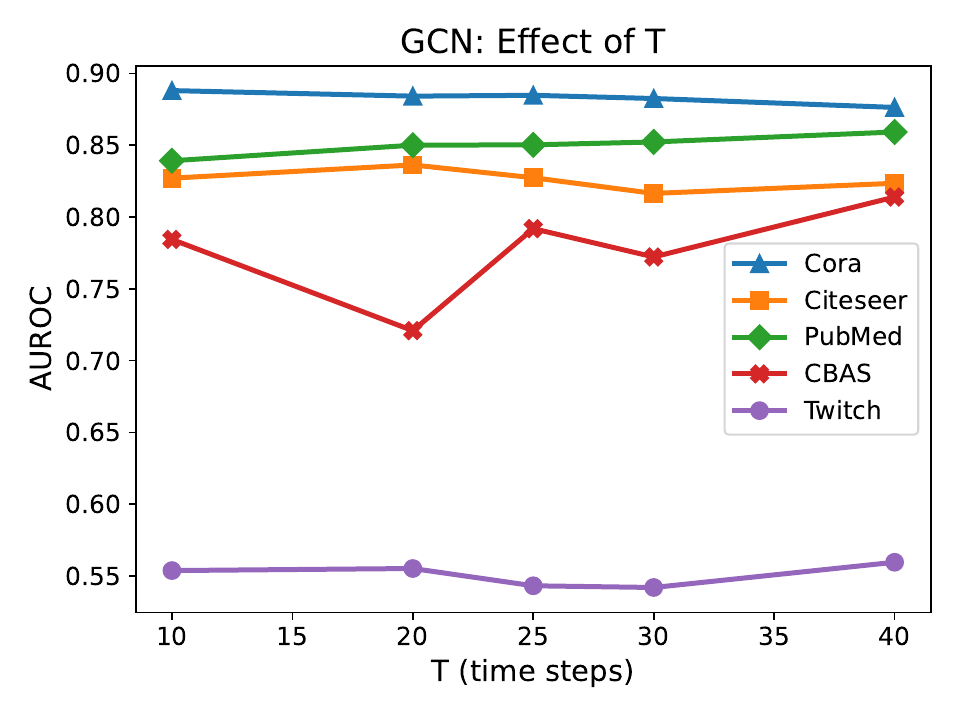}
        \caption{$\text{GCN}_T$}
        \label{fig:sub2}
    \end{subfigure}\hfill
    \begin{subfigure}{0.31\textwidth}
        \centering
        \includegraphics[width=\linewidth]{ 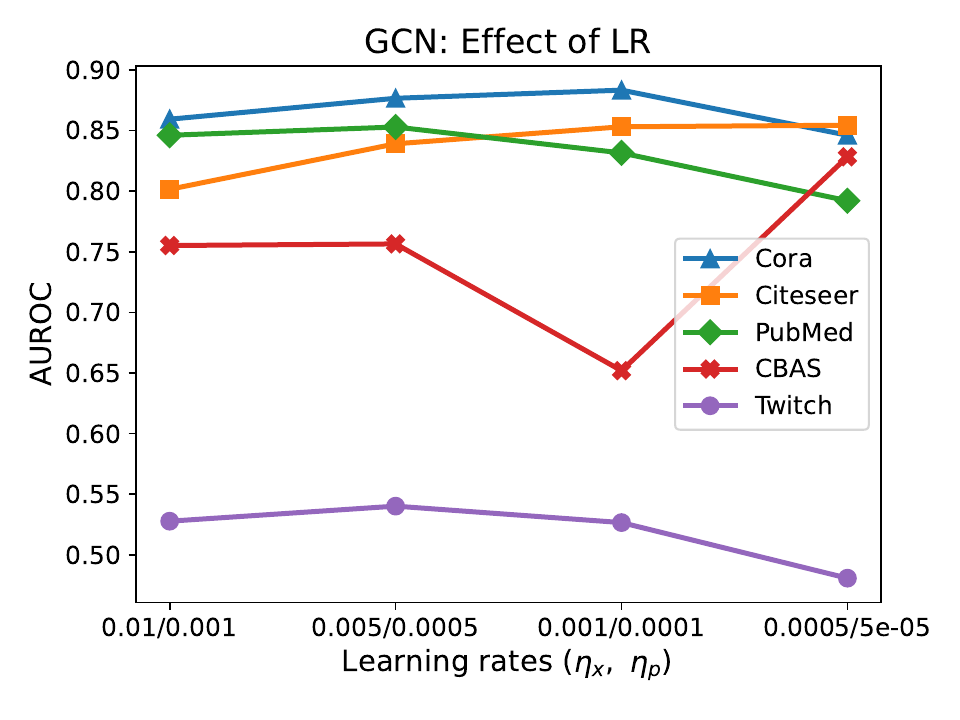}
        \caption{$\text{GCN}_{LR}$}
        \label{fig:sub3}
    \end{subfigure}
    
    \vspace{0.6em}
    
    \begin{subfigure}{0.31\textwidth}
        \centering
        \includegraphics[width=\linewidth]{ 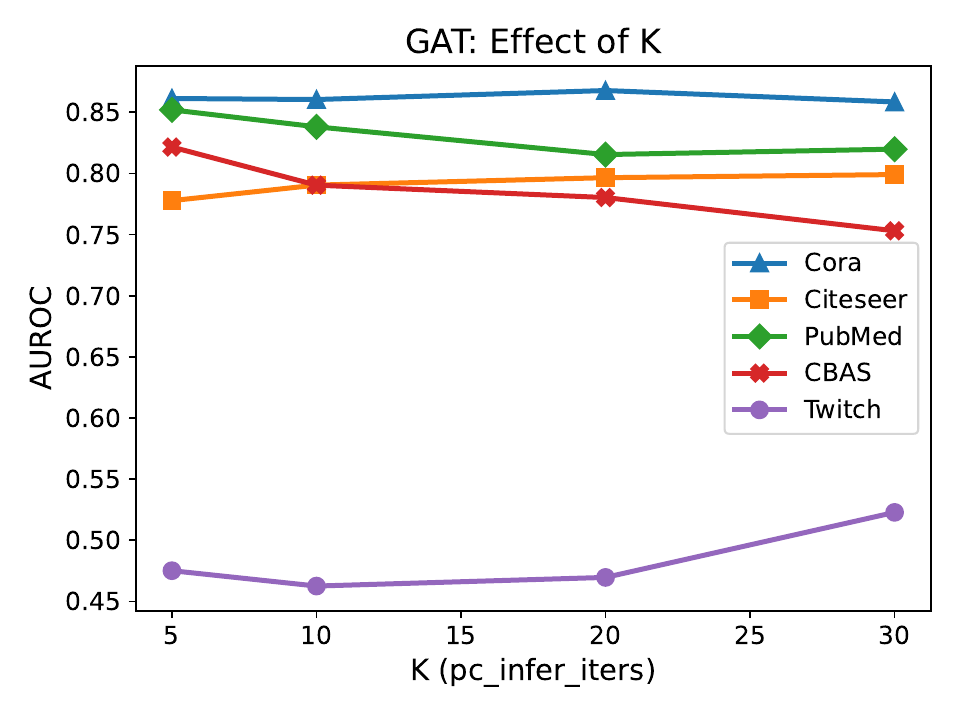}
        \caption{$\text{GAT}_K$}
        \label{fig:sub4}
    \end{subfigure}\hfill
    \begin{subfigure}{0.31\textwidth}
        \centering
        \includegraphics[width=\linewidth]{ 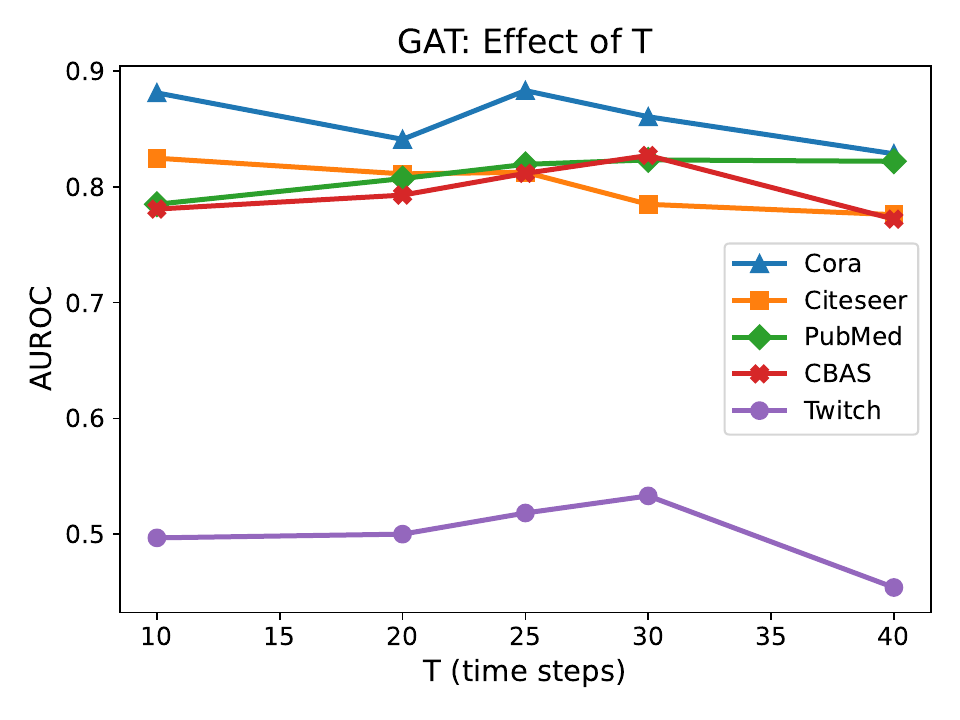}
        \caption{$\text{GAT}_T$}
        \label{fig:sub5}
    \end{subfigure}\hfill
    \begin{subfigure}{0.31\textwidth}
        \centering 
        \includegraphics[width=\linewidth]{ 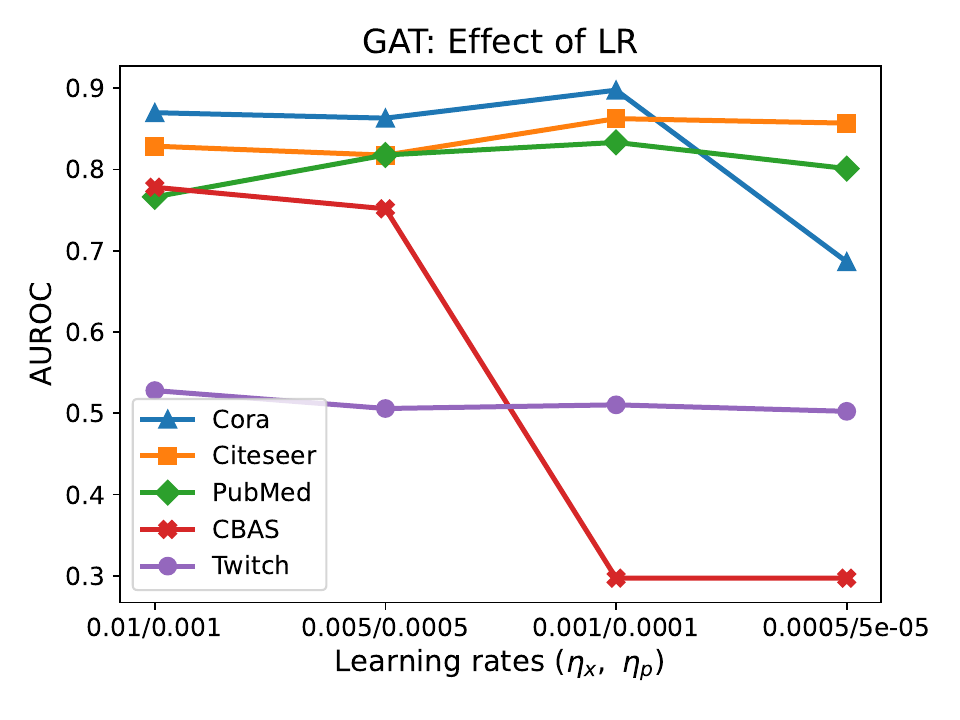}
        \caption{$\text{GAT}_{LR}$}
        \label{fig:sub6}
    \end{subfigure}
    \caption{Sensitivity analysis of SGPC under predictive coding iterations ($K$), timesteps ($T$), and learning rate ($LR$) on AUROC are evaluated across five datasets.}
    \label{fig:sensitivity}
\end{figure*}

\subsection{Additional Results}
\label{app:results}

\subsubsection{Comparison of Performance and Calibration}
\label{app:Calibration}
Table~\ref{tab:PerformanceGAT} reports node classification accuracy and calibration metrics under both ID and OOD settings with the GAT backbone. Overall, SIGHT consistently delivers competitive or superior performance compared to GAT and G-$\Delta$UQ. Apart from CBAS, SIGHT achieves strong OOD accuracy while simultaneously reducing calibration errors, as reflected in the lowest OOD ECE and Brier scores. These improvements highlight the ability of predictive coding residuals to enhance uncertainty awareness beyond conventional GAT models. On the smaller CBAS datasets, SIGHT remains reliable, matching or surpassing baselines in ID accuracy and achieving higher AUROC. Importantly, AUROC values confirm that SIGHT provides more reliable uncertainty estimation under distribution shifts, establishing a favorable balance between predictive accuracy and calibration quality across diverse domains.

\subsubsection{Explainability of Predictive Coding on Uncertainty Estimation}
\label{app:Explainability}
Other than OOD data shown in Figure~\ref{fig:pc}, Figure~\ref{fig:pcID3} reports the correlation between PC error and confidence on the ID test sets of Cora, Citeseer, and Pubmed. Similar to the OOD case, we observe a strong negative correlation: higher PC errors correspond to lower confidence. This shows that predictive coding residuals not only capture local mismatches between predictions and inputs but also faithfully reflect model certainty on in-distribution samples. The high $R^2$ values of the polynomial fits further confirm that PC errors provide reliable, interpretable indicators of confidence, supporting the claim that SIGHT yields uncertainty estimates without post-hoc calibration.

\subsubsection{Parameter Analysis}
\label{app:Parameter}
As introduced in~\ref{sec:exre}, we further investigate the impact of different hyperparameters on SIGHT, including the number of predictive coding iterations $K$, the number of timesteps $T$, and the learning rate under both GCN and GAT backbones. Figure~\ref{fig:sensitivity} presents the sensitivity analysis. Across datasets, performance remains stable across a wide range of $K$ and $T$, showing that SIGHT is robust to the choice of inference iterations and simulation length. In contrast, the learning rate has a more pronounced influence: overly large or small values lead to noticeable drops in AUROC, particularly on smaller datasets such as CBAS and Twitch. These findings demonstrate that SIGHT maintains strong and consistent performance without requiring fine-grained hyperparameter tuning, highlighting its practicality and trustworthiness in real-world settings.

\end{document}